\pdfoutput=1

\documentclass[11pt]{article}

\usepackage{naacl2021}

\usepackage{times}
\usepackage{latexsym}

\usepackage[T1]{fontenc}

\usepackage[utf8]{inputenc}

\usepackage{microtype}

\usepackage{xspace}
\usepackage{multirow}

\usepackage{natbib}
\usepackage{paralist}

\usepackage{array}
\newcolumntype{x}[1]{>{\centering\arraybackslash\hspace{0pt}}p{#1}}

\usepackage{times}

\usepackage{soul}
\usepackage{url}
\usepackage[utf8]{inputenc}
\usepackage{caption}
\usepackage{graphicx}
\usepackage{amsmath}
\usepackage{booktabs}
\usepackage{enumitem}
\usepackage{microtype}
\usepackage{booktabs}

\usepackage{pgf}
\usepackage{pgfkeys}
\usepackage{pgfplots,pgfplotstable}
\usepackage{tikz}
\usetikzlibrary{3d}
\usetikzlibrary{arrows.meta}
\usetikzlibrary{backgrounds}
\usetikzlibrary{bayesnet}
\usetikzlibrary{calc}
\usetikzlibrary{calligraphy}
\usetikzlibrary{fit}
\usetikzlibrary{positioning}
\usetikzlibrary{decorations.markings}
\usetikzlibrary{decorations.pathreplacing}
\usetikzlibrary{shapes}
\usetikzlibrary{shapes.geometric}
\usetikzlibrary{shapes.multipart}
\usetikzlibrary{tikzmark}
\usetikzlibrary{shapes}
\usetikzlibrary{shapes.geometric}

\definecolor{saffron}{RGB}{234, 196, 53}
\definecolor{teal}{RGB}{0, 123, 159}
\definecolor{aquamarine}{RGB}{51, 153, 255}

\newcommand{\eg}{e.g.,\xspace}

\newcommand{\ie}{i.e.,\xspace}

\makeatother

\newcommand\Tstrut{\rule{0pt}{2.6ex}}

\definecolor{Red}{rgb}{1,0,0}
\definecolor{Green}{rgb}{0,0.69,0}
\definecolor{Blue}{rgb}{0,0,1}
\definecolor{LightBlue}{rgb}{0,0.5,1}
\definecolor{veryLightBlue}{rgb}{0.85,0.98,1}
\definecolor{veryLightGreen}{rgb}{0.6,1,0.6}
\definecolor{Skin}{rgb}{1,0.71,0.69}
\definecolor{Grey}{rgb}{0.5,0.5,0.5}
\definecolor{LightGrey}{rgb}{0.6,0.6,0.6}
\definecolor{Black}{rgb}{0,0,0}
\definecolor{White}{rgb}{1,1,1}
\definecolor{brickred}{rgb}{0.8, 0.25, 0.33}

\newcommand{\green}{\color{Green}}
\newcommand{\blue}{\color{Blue}}

\usepackage{datatool}

\makeatletter
\providecommand*{\xdtlgetrowindex}[4]{%
  \protected@edef\dtl@dogetrowindex{\noexpand\@dtlgetrowindex{\noexpand#1}{#2}{\number#3}{#4}%
  \dtl@dogetrowindex
}}
\makeatother

\DTLnewdb{gender}
\DTLnewrow{gender}
\DTLnewdbentry{gender}{symbol}{weat6}
\DTLnewdbentry{gender}{name}{C6: M/W, Career/Fam}
\DTLnewdbentry{gender}{shortname}{C6}
\DTLnewrow{gender}
\DTLnewdbentry{gender}{symbol}{weat8}
\DTLnewdbentry{gender}{name}{C8: Science/Arts, M/W}
\DTLnewdbentry{gender}{shortname}{C8}
\DTLnewrow{gender}
\DTLnewdbentry{gender}{symbol}{weat11}
\DTLnewdbentry{gender}{name}{C11: M/W, Pleasant}
\DTLnewdbentry{gender}{shortname}{C11}
\DTLnewrow{gender}
\DTLnewdbentry{gender}{symbol}{heilman_MF_double_bind_competent}
\DTLnewdbentry{gender}{name}{Competent: M/W, Competent}
\DTLnewdbentry{gender}{shortname}{Competent}
\DTLnewrow{gender}
\DTLnewdbentry{gender}{symbol}{heilman_MF_double_bind_likable}
\DTLnewdbentry{gender}{name}{Likeable: M/W, Likeable}
\DTLnewdbentry{gender}{shortname}{Likeable} 
\DTLnewrow{gender}
\DTLnewdbentry{gender}{symbol}{occ_gender}
\DTLnewdbentry{gender}{name}{Occupation: M/W, Occupation}
\DTLnewdbentry{gender}{shortname}{Occupation}

\DTLnewdb{race}
\DTLnewrow{race}
\DTLnewdbentry{race}{symbol}{weat3}
\DTLnewdbentry{race}{name}{C3: EA/AA, Pleasant}
\DTLnewdbentry{race}{shortname}{C3}
\DTLnewrow{race}
\DTLnewdbentry{race}{symbol}{weat12}
\DTLnewdbentry{race}{name}{C12: EA/AA, Career/Family}
\DTLnewdbentry{race}{shortname}{C12}
\DTLnewrow{race}
\DTLnewdbentry{race}{symbol}{weat13}
\DTLnewdbentry{race}{name}{C13: EA/AA, Science/Arts}
\DTLnewdbentry{race}{shortname}{C13}
\DTLnewrow{race}
\DTLnewdbentry{race}{symbol}{heilman_race_double_bind_competent}
\DTLnewdbentry{race}{name}{Double Bind: EA/AA, Competent}
\DTLnewdbentry{race}{shortname}{Competent}
\DTLnewrow{race}
\DTLnewdbentry{race}{symbol}{heilman_race_double_bind_likable}
\DTLnewdbentry{race}{name}{Double Bind: EA/AA, Likeable}
\DTLnewdbentry{race}{shortname}{Likeable}
\DTLnewrow{race}
\DTLnewdbentry{race}{symbol}{occ_race}
\DTLnewdbentry{race}{name}{Occupation: EA/AA, Occupation}
\DTLnewdbentry{race}{shortname}{Occupation}
\DTLnewrow{race}
\DTLnewdbentry{race}{symbol}{angry_race}
\DTLnewdbentry{race}{name}{Angry Black Woman Stereotype}
\DTLnewdbentry{race}{shortname}{ABW}

\DTLnewdb{genderC}
\DTLnewrow{genderC}
\DTLnewdbentry{genderC}{symbol}{weat6}
\DTLnewdbentry{genderC}{name}{C6: M/W, Career/Family}
\DTLnewdbentry{genderC}{shortname}{C6}
\DTLnewrow{genderC}
\DTLnewdbentry{genderC}{symbol}{occ_gender}
\DTLnewdbentry{genderC}{name}{Occupation: M/W, Occupation}
\DTLnewdbentry{genderC}{shortname}{Occupation}

\DTLloaddb{visualbert_exp1_google}{visualbert/exp1_google.csv}
\DTLloaddb{visualbert_exp2_google}{visualbert/exp2_google.csv}
\DTLloaddb{visualbert_exp3a_google}{visualbert/exp4a_google.csv}
\DTLloaddb{visualbert_exp3b_google}{visualbert/exp4b_google.csv}
\DTLloaddb{visualbert_exp1_coco}{visualbert/exp1_coco.csv}
\DTLloaddb{visualbert_exp2_coco}{visualbert/exp2_coco.csv}
\DTLloaddb{visualbert_exp3a_coco}{visualbert/exp4a_coco.csv}
\DTLloaddb{vilbert_exp1_google}{vilbert/exp1_google.csv}
\DTLloaddb{vilbert_exp2_google}{vilbert/exp2_google.csv}
\DTLloaddb{vilbert_exp3a_google}{vilbert/exp4a_google.csv}
\DTLloaddb{vilbert_exp3b_google}{vilbert/exp4b_google.csv}
\DTLloaddb{vilbert_exp1_concap}{vilbert/exp1_concap.csv}
\DTLloaddb{vilbert_exp2_concap}{vilbert/exp2_concap.csv}
\DTLloaddb{vilbert_exp3a_concap}{vilbert/exp4a_concap.csv}
\DTLloaddb{vilbert_exp3b_concap}{vilbert/exp4b_concap.csv}
\DTLloaddb{lxmert_exp1_google}{lxmert/exp1_google.csv}
\DTLloaddb{lxmert_exp2_google}{lxmert/exp2_google.csv}
\DTLloaddb{lxmert_exp3a_google}{lxmert/exp4a_google.csv}
\DTLloaddb{lxmert_exp3b_google}{lxmert/exp4b_google.csv}
\DTLloaddb{lxmert_exp1_coco}{lxmert/exp1_coco.csv}
\DTLloaddb{lxmert_exp2_coco}{lxmert/exp2_coco.csv}
\DTLloaddb{lxmert_exp3a_coco}{lxmert/exp4a_coco.csv}
\DTLloaddb{lxmert_exp3b_coco}{lxmert/exp4b_coco.csv}
\DTLloaddb{vlbert_exp1_google}{vlbert/exp1_google.csv}
\DTLloaddb{vlbert_exp2_google}{vlbert/exp2_google.csv}
\DTLloaddb{vlbert_exp3a_google}{vlbert/exp4a_google.csv}
\DTLloaddb{vlbert_exp3b_google}{vlbert/exp4b_google.csv}
\DTLloaddb{vlbert_exp1_concap}{vlbert/exp1_concap.csv}
\DTLloaddb{vlbert_exp2_concap}{vlbert/exp2_concap.csv}
\DTLloaddb{vlbert_exp3a_concap}{vlbert/exp4a_concap.csv}
\DTLloaddb{vlbert_exp3b_concap}{vlbert/exp4b_concap.csv}

\newcommand{\DTLfetchRound}[5]{%
  \xdtlgetrowindex{\myrowidx}{#1}{\dtlcolumnindex{#1}{#2}}{#3}%
  \ifx\myrowidx\dtlnovalue%
    {\hspace{2ex}--}%
  \else%
    \edtlgetrowforvalue{#1}{\dtlcolumnindex{#1}{#2}}{#3}%
    \dtlgetentryfromcurrentrow{\dtlcurrentvalue}{\dtlcolumnindex{#1}{#5}}%
    \DTLifnumgt{0.05}{\dtlcurrentvalue}{\leavevmode\color{blue}\itshape}{}%
    \dtlgetentryfromcurrentrow{\dtlcurrentvalue}{\dtlcolumnindex{#1}{#4}}%
    \DTLround{\dtlcurrentvalue}{\dtlcurrentvalue}{2}%
    \dtlcurrentvalue%
  \fi%
}

\newcommand{\showExperiment}[1]{%
  \centering\small\makebox[0pt][c]{%
\raisebox{0ex}{\begin{minipage}[t]{0.57\textwidth}
\begin{tabular*}{\textwidth}{@{\extracolsep{\fill}}p{12.5em}p{0.5em}*{4}{|>{\raggedleft\arraybackslash}p{2.2em}}}
  \textbf{Gender}
  & \rotatebox{90}{\scalebox{0.8}{\parbox{4em}{\centering\footnotesize Level}}}
  & \hspace*{-0.8ex}\rotatebox{60}{\scalebox{0.8}{\parbox{4em}{\centering\footnotesize VisualBERT Google}}}
  & \hspace*{-0.8ex}\rotatebox{60}{\scalebox{0.8}{\parbox{4em}{\centering\footnotesize ViLBERT Google}}}
  & \hspace*{-0.8ex}\rotatebox{60}{\scalebox{0.8}{\parbox{4em}{\centering\footnotesize 
  LXMert Google}}}%
  & \hspace*{-0.8ex}\rotatebox{60}{\scalebox{0.8}{\parbox{4em}{\centering\footnotesize
  VLBERT Google}}}%
    \vspace{-2ex}%
    \DTLforeach{gender}%
    {\symbol=symbol,\name=name}%
    {%
  \DTLiffirstrow{\\}{\\\hline}%
  \Tstrut\multirow{3}{*}{\name}%
  & W%
  & \DTLfetchRound{visualbert_#1_google}{test_name}{\symbol:word}{esize}{pval}%
  & \DTLfetchRound{vilbert_#1_google}{test_name}{\symbol:word}{esize}{pval}%
  & \DTLfetchRound{lxmert_#1_google}{test_name}{\symbol:word}{esize}{pval}%
  & \DTLfetchRound{vlbert_#1_google}{test_name}{\symbol:word}{esize}{pval}%
  \\%
  & {\leavevmode\hspace*{0.4ex}S}%
  & \DTLfetchRound{visualbert_#1_google}{test_name}{\symbol:sent}{esize}{pval}%
  & \DTLfetchRound{vilbert_#1_google}{test_name}{\symbol:sent}{esize}{pval}%
  & \DTLfetchRound{lxmert_#1_google}{test_name}{\symbol:sent}{esize}{pval}%
  & \DTLfetchRound{vlbert_#1_google}{test_name}{\symbol:sent}{esize}{pval}%
  \\%
  & {\leavevmode\hspace*{0.2ex}C}%
  & \DTLfetchRound{visualbert_#1_google}{test_name}{\symbol:contextual}{esize}{pval}%
  & \DTLfetchRound{vilbert_#1_google}{test_name}{\symbol:contextual}{esize}{pval}%
  & \DTLfetchRound{lxmert_#1_google}{test_name}{\symbol:contextual}{esize}{pval}%
  & \DTLfetchRound{vlbert_#1_google}{test_name}{\symbol:contextual}{esize}{pval}%
    }\\
\end{tabular*}%
\end{minipage}}%
\hspace*{2em}\begin{minipage}[t]{0.60\textwidth}
  \begin{tabular*}{\linewidth}{@{\extracolsep{\fill}}p{14em}p{0.5em}*{4}{|>{\raggedleft\arraybackslash}p{2.2em}}}%
  \textbf{Race}
  & \rotatebox{90}{\scalebox{0.8}{\parbox{4em}{\centering\footnotesize Level}}}
  & \hspace*{-0.8ex}\rotatebox{60}{\scalebox{0.8}{\parbox{4em}{\centering\footnotesize VisualBERT Google}}}
  & \hspace*{-0.8ex}\rotatebox{60}{\scalebox{0.8}{\parbox{4em}{\centering\footnotesize ViLBERT Google}}}
  & \hspace*{-0.8ex}\rotatebox{60}{\scalebox{0.8}{\parbox{4em}{\centering\footnotesize LXMert Google}}}
  & \hspace*{-0.8ex}\rotatebox{60}{\scalebox{0.8}{\parbox{4em}{\centering\footnotesize VLBERT Google}}}
    \vspace{-3ex}%
    \DTLforeach{race}%
    {\symbol=symbol,\name=name}%
    {%
  \DTLiffirstrow{\\}{\\\hline}%
  \Tstrut\multirow{3}{*}{\name}%
  & W%
  & \DTLfetchRound{visualbert_#1_google}{test_name}{\symbol:word}{esize}{pval}%
  & \DTLfetchRound{vilbert_#1_google}{test_name}{\symbol:word}{esize}{pval}%
  & \DTLfetchRound{lxmert_#1_google}{test_name}{\symbol:word}{esize}{pval}%
  & \DTLfetchRound{vlbert_#1_google}{test_name}{\symbol:word}{esize}{pval}%
  \\%
  & {\leavevmode\hspace*{0.4ex}S}%
  & \DTLfetchRound{visualbert_#1_google}{test_name}{\symbol:sent}{esize}{pval}%
  & \DTLfetchRound{vilbert_#1_google}{test_name}{\symbol:sent}{esize}{pval}%
  & \DTLfetchRound{lxmert_#1_google}{test_name}{\symbol:sent}{esize}{pval}%
  & \DTLfetchRound{vlbert_#1_google}{test_name}{\symbol:sent}{esize}{pval}%
  \\%
  & {\leavevmode\hspace*{0.2ex}C}%
  & \DTLfetchRound{visualbert_#1_google}{test_name}{\symbol:contextual}{esize}{pval}%
  & \DTLfetchRound{vilbert_#1_google}{test_name}{\symbol:contextual}{esize}{pval}%
  & \DTLfetchRound{lxmert_#1_google}{test_name}{\symbol:contextual}{esize}{pval}%
  & \DTLfetchRound{vlbert_#1_google}{test_name}{\symbol:contextual}{esize}{pval}%
    }\\
    \end{tabular*}
\end{minipage}}
}

\usepackage{mathtools}
\usepackage{cleveref}

\DeclareMathOperator*{\mean}{mean}
\DeclareMathOperator*{\stddev}{std\_dev}

\title{Measuring Social Biases in Grounded Vision and Language Embeddings}

\author{
  Candace Ross, Boris Katz \& Andrei Barbu \\
  CSAIL, Massachusetts Institute of Technology \\
  \texttt{\{ccross,boris,abarbu\}@mit.edu}
}

\begin{document}
\begin{filecontents*}{visualbert/exp1_google.csv}
test_name,esize,pval
weat3:word,0.23010873924402506,0.00161
weat3:sent,0.3083328769191792,0.00001
weat3:contextual,-0.0074982435274964085,0.54798
weat6:word,0.5715717376727245,0.00015
weat6:sent,-0.18169449701466803,0.99946
weat6:contextual,-0.609012840721026,1.0
weat8:word,0.7690598008999587,0.00001
weat8:sent,0.6164188777774718,0.00001
weat8:contextual,0.2968849296786705,0.00001
weat11:word,-0.6581512354176247,0.99999
weat11:sent,-0.7406719188079499,1.0
weat11:contextual,0.42299111541053713,0.00001
weat12:word,-0.2922141390276468,0.9999
weat12:sent,-0.5351629836433123,1.0
weat12:contextual,0.3625615351270848,0.00001
weat13:word,0.04306141851649377,0.29269
weat13:sent,0.11832623895311063,0.02866
weat13:contextual,0.5803013269198949,0.00001
heilman_MF_double_bind_competent:sent,-0.28015326441244337,0.96231
heilman_MF_double_bind_competent:contextual,-0.6719337427954373,1.0
heilman_MF_double_bind_likable:sent,0.09589013942805712,0.2743
heilman_MF_double_bind_likable:contextual,-0.4189787405661641,0.99588
heilman_MF_double_bind_competent:word,-0.23488944232644426,0.93108
heilman_MF_double_bind_likable:word,-1.2410303155105362,1.0
heilman_race_double_bind_competent:word,0.7499673594608379,0.00001
heilman_race_double_bind_competent:sent,0.9972912601857249,0.00001
heilman_race_double_bind_competent:contextual,1.1044029189774407,0.00001
heilman_race_double_bind_likable:word,-0.2455255301025526,0.93928
heilman_race_double_bind_likable:sent,-0.08831565232274217,0.71072
heilman_race_double_bind_likable:contextual,0.9749052914526616,0.00001
occ_gender:word,0.017745298543821238,0.45886
occ_gender:sent,0.7717280919278335,0.00001
occ_gender:contextual,0.976640341411622,0.00001
angry_race:word,-0.0652853167836462,0.59847
angry_race:sent,-0.5022203688393977,1.0
angry_race:contextual,0.7140197585816661,0.00001
occ_race:word,-0.14963211220063077,0.94947
occ_race:sent,-0.2602123592725174,0.92957
occ_race:contextual,-0.7045896738825178,0.99995
\end{filecontents*}

\begin{filecontents*}{visualbert/exp2_google.csv}
test_name,esize,pval
weat3:word,1.549792093989554,0.00001
weat3:sent,1.5377441613113787,0.00001
weat3:contextual,0.26170717554273487,0.00002
weat6:word,1.0522921583309706,0.00001
weat6:sent,-0.5692875304517995,1.0
weat6:contextual,-0.8633433121151375,1.0
weat8:word,0.7690598008999587,0.00001
weat8:sent,0.6164188777774718,0.00001
weat8:contextual,0.2968849296786705,0.00001
weat11:word,-1.4821245342477316,1.0
weat11:sent,-1.1290393163011434,1.0
weat11:contextual,-0.15036279870493383,0.99644
weat12:word,-0.03724574070091007,0.6829
weat12:sent,0.35827432115451124,0.00001
weat12:contextual,0.8357752914260482,0.00001
weat13:word,-1.7366260153731237,1.0
weat13:sent,-0.08037511051921611,0.90053
weat13:contextual,1.0019502454481664,0.00001
heilman_MF_double_bind_competent:sent,-0.11820454277604693,0.77218
heilman_MF_double_bind_competent:contextual,-0.5963664712441632,0.99993
heilman_MF_double_bind_likable:sent,1.763877954512846,0.00001
heilman_MF_double_bind_likable:contextual,-0.10736418278367735,0.74971
heilman_MF_double_bind_competent:word,0.22994687235013472,0.07237
heilman_MF_double_bind_likable:word,-1.3072406069301978,1.0
heilman_race_double_bind_competent:word,1.1309463116710152,0.00001
heilman_race_double_bind_competent:sent,1.2495145206987215,0.00001
heilman_race_double_bind_competent:contextual,1.1133112237028067,0.00001
heilman_race_double_bind_likable:word,0.2905970998595459,0.03286
heilman_race_double_bind_likable:sent,0.4160012754108181,0.00408
heilman_race_double_bind_likable:contextual,0.9251681220008352,0.00001
occ_gender:word,-0.7740761670691733,1.0
occ_gender:sent,0.32956880251763193,0.0313
occ_gender:contextual,0.8984612987069002,0.00001
angry_race:word,0.3427751785185533,0.09281
angry_race:sent,0.4872618537775402,0.00001
angry_race:contextual,1.7099548270436926,0.00001
occ_race:word,-0.03574049665832521,0.65127
occ_race:sent,0.14645079114919338,0.20578
occ_race:contextual,-0.5745466769721378,0.9995
\end{filecontents*}

\begin{filecontents*}{visualbert/exp4a_google.csv}
test_name,esize,pval
weat3:mask_t_word,0.3700735134281056,0.00001
weat3:mask_t_sent,0.32935324203636573,0.00001
weat6:mask_t_word,0.2176674018904672,0.08459
weat6:mask_t_sent,0.14255857870454167,0.00536
weat8:mask_t_word,0.5885300309656879,0.00016
weat8:mask_t_sent,0.4589219421937157,0.00001
weat11:mask_t_word,-0.3673015110617337,0.98988
weat11:mask_t_sent,-0.46912902307976584,1.0
weat12:mask_t_word,-0.5468076632866667,1.0
weat12:mask_t_sent,-0.5217956620279407,1.0
weat13:mask_t_word,-0.39863490040899546,1.0
weat13:mask_t_sent,-0.001944400538332134,0.51079
heilman_MF_double_bind_competent:mask_t_sent,-0.061852746469723864,0.65168
heilman_MF_double_bind_likable:mask_t_sent,-0.07470832996544675,0.6803
heilman_MF_double_bind_competent:mask_t_word,-0.0023269206385038163,0.50562
heilman_MF_double_bind_likable:mask_t_word,-0.12038222579213798,0.77602
heilman_race_double_bind_competent:mask_t_word,1.0202935230087973,0.00001
heilman_race_double_bind_competent:mask_t_sent,-0.4352532142589058,0.99729
heilman_race_double_bind_likable:mask_t_word,0.32479182880831386,0.01993
heilman_race_double_bind_likable:mask_t_sent,-0.6830952245767857,1.0
occ_gender:mask_t_word,0.22573237982617034,0.0773
occ_gender:mask_t_sent,0.048298321343883455,0.39405
angry_race:mask_t_word,0.3608874223622063,0.08098
angry_race:mask_t_sent,0.7609303311492867,0.00001
occ_race:mask_t_word,-0.6255860012898701,1.0
occ_race:mask_t_sent,-0.26762861304392854,0.93423
\end{filecontents*}

\begin{filecontents*}{visualbert/exp4b_google.csv}
test_name,esize,pval
\end{filecontents*}

\begin{filecontents*}{visualbert/exp1_coco.csv}
test_name,esize,pval
occ_gender:word,-0.0709499969449513,0.6697
occ_gender:sent,-0.227340838302158,0.96635
occ_gender:contextual,-0.31570426437040994,0.99464
weat6:word,0.1302543296420215,0.20802
weat6:sent,0.2832913962054409,0.01455
weat6:contextual,-0.20061587991522517,0.93736
weat8:word,0.31457223014505376,0.02269
weat8:sent,0.49110418996162697,0.00395
weat8:contextual,0.4917399319443303,0.00401
\end{filecontents*}

\begin{filecontents*}{visualbert/exp2_coco.csv}
test_name,esize,pval
occ_gender:word,-0.6449436425468368,1.0
occ_gender:sent,0.08996999324323862,0.23385
occ_gender:contextual,-0.3462146707738554,0.99731
weat6:word,0.1516334840099042,0.16998
weat6:sent,0.4113268071910274,0.00081
weat6:contextual,-0.9889113636317216,1.0
weat8:word,0.31457223014505376,0.02245
weat8:sent,0.49110418996162697,0.00408
weat8:contextual,0.4917399319443303,0.00391
\end{filecontents*}

\begin{filecontents*}{visualbert/exp4a_coco.csv}
test_name,esize,pval
\end{filecontents*}

\begin{filecontents*}{vilbert/exp1_google.csv}
test_name,esize,pval
weat3:word,0.3109775001363372,0.00003
weat3:sent,0.2526955509747277,0.00003
weat3:contextual,-0.2932400709143942,1.0
weat6:word,1.0372670978707215,0.00001
weat6:sent,0.9838898634117222,0.00001
weat6:contextual,0.7614299284169891,0.00001
weat8:word,0.5931207758208913,0.00032
weat8:sent,0.26474220090023454,0.00001
weat8:contextual,-0.31544738218152907,1.0
weat11:word,-0.9066504822503844,1.0
weat11:sent,-1.0791858948110233,1.0
weat11:contextual,-0.6167245220690801,1.0
weat12:word,0.04492074980879466,0.28841
weat12:sent,0.05241854551135918,0.20003
weat12:contextual,0.9246189027957946,0.00001
weat13:word,0.6125253177039743,0.00001
weat13:sent,0.34775661414721937,0.00001
weat13:contextual,1.0895018507858891,0.00001
heilman_MF_double_bind_competent:word,-0.5749912943749903,0.99944
heilman_MF_double_bind_competent:sent,-0.29395791305709745,0.95247
heilman_MF_double_bind_competent:contextual,0.1953631139877187,0.1354
heilman_MF_double_bind_likable:word,-1.2559244345950407,1.0
heilman_MF_double_bind_likable:sent,-0.12323247376726457,0.75567
heilman_MF_double_bind_likable:contextual,1.2516794566850888,0.00001
heilman_race_double_bind_competent:word,1.280779150531661,0.00001
heilman_race_double_bind_competent:sent,1.1361581935969003,0.00001
heilman_race_double_bind_competent:contextual,1.18565050997875,0.00001
heilman_race_double_bind_likable:word,0.4147617648807821,0.00946
heilman_race_double_bind_likable:sent,0.7253356466398725,0.00002
heilman_race_double_bind_likable:contextual,1.0893274385584202,0.00001
angry_race:word,0.4148430722145216,0.05367
angry_race:sent,0.4559797207748109,0.00001
angry_race:contextual,0.6561696236213967,0.00001
occ_gender:word,0.8578515984090754,0.00001
occ_gender:sent,0.9470066375683872,0.00001
occ_gender:contextual,1.531418431880717,0.00001
occ_race:word,-0.4146233565594685,0.99998
occ_race:sent,-0.2582101907438492,0.92821
occ_race:contextual,-0.3725367277518173,0.98259
\end{filecontents*}

\begin{filecontents*}{vilbert/exp2_google.csv}
test_name,esize,pval
weat3:word,1.0332541455966682,0.00001
weat3:sent,0.8534035822727479,0.00001
weat3:contextual,-0.14039089316549147,0.98729
weat6:word,1.0919697375685133,0.00001
weat6:sent,1.340662281675401,0.00001
weat6:contextual,0.6543893448020233,0.00001
weat8:word,0.5931207758208913,0.00028
weat8:sent,0.26474220090023454,0.00003
weat8:contextual,-0.31544738218152907,1.0
weat11:word,-1.3293768641879,1.0
weat11:sent,-1.1703382053449676,1.0
weat11:contextual,-0.46456926432335677,1.0
weat12:word,0.8826015235127818,0.00001
weat12:sent,0.8106545269216816,0.00001
weat12:contextual,1.0168095378371094,0.00001
weat13:word,1.2734179494474076,0.00001
weat13:sent,1.044372080515638,0.00001
weat13:contextual,1.388691464115408,0.00001
heilman_MF_double_bind_competent:word,0.23459589103160478,0.09208
heilman_MF_double_bind_competent:sent,-0.3473577511289945,0.9746
heilman_MF_double_bind_competent:contextual,-0.08378642064310955,0.68046
heilman_MF_double_bind_likable:word,-0.607430816516019,0.99976
heilman_MF_double_bind_likable:sent,-0.1632456168332632,0.82109
heilman_MF_double_bind_likable:contextual,1.3052106975373363,0.00001
heilman_race_double_bind_competent:word,1.557196127043254,0.00001
heilman_race_double_bind_competent:sent,1.4509795989506291,0.00001
heilman_race_double_bind_competent:contextual,1.2011924869437045,0.00001
heilman_race_double_bind_likable:word,1.127801346086892,0.00001
heilman_race_double_bind_likable:sent,1.0416681350179147,0.00001
heilman_race_double_bind_likable:contextual,1.1159522633381505,0.00001
angry_race:word,-0.2757910272186856,0.85432
angry_race:sent,-0.5254184602227959,1.0
angry_race:contextual,1.4400443670909133,0.00001
occ_gender:word,0.04709970011316998,0.39541
occ_gender:sent,0.2170117325811144,0.11097
occ_gender:contextual,1.458785545338315,0.00001
occ_race:word,-0.48289927682976796,1.0
occ_race:sent,-0.17893795546367924,0.84378
occ_race:contextual,-0.1887906649593474,0.85405
\end{filecontents*}

\begin{filecontents*}{vilbert/exp4a_google.csv}
test_name,esize,pval
weat3:mask_t_word,0.25374389783946893,0.00063
weat3:mask_t_sent,0.34443956912517754,0.00001
weat6:mask_t_word,1.1255834396913724,0.00001
weat6:mask_t_sent,1.0010873343293856,0.00001
weat8:mask_t_word,0.8671583889513026,0.00001
weat8:mask_t_sent,0.40997737610876644,0.00001
weat11:mask_t_word,-1.1867448496686384,1.0
weat11:mask_t_sent,-1.2062519822081346,1.0
weat12:mask_t_word,-0.06478171251169998,0.79434
weat12:mask_t_sent,0.045163967783918256,0.23525
weat13:mask_t_word,0.4640194760873118,0.00001
weat13:mask_t_sent,0.3328107189965423,0.00001
heilman_MF_double_bind_competent:mask_t_word,-0.5217795371626738,0.99851
heilman_MF_double_bind_competent:mask_t_sent,-0.3966446152939426,0.98746
heilman_MF_double_bind_likable:mask_t_word,-1.4057479802554582,1.0
heilman_MF_double_bind_likable:mask_t_sent,-0.1812659165145427,0.84612
heilman_race_double_bind_competent:mask_t_word,1.2444746069120858,0.00001
heilman_race_double_bind_competent:mask_t_sent,1.1039570862719097,0.00001
heilman_race_double_bind_likable:mask_t_word,0.6440074381120875,0.0001
heilman_race_double_bind_likable:mask_t_sent,0.5767958731969475,0.00048
angry_race:mask_t_word,0.438679550105447,0.0459
angry_race:mask_t_sent,0.5374818160015792,0.00001
occ_gender:mask_t_word,1.0419260162761153,0.00001
occ_gender:mask_t_sent,1.076857985930171,0.00001
occ_race:mask_t_word,-0.26544621026143433,0.99576
occ_race:mask_t_sent,-0.24388096913032964,0.91601
\end{filecontents*}

\begin{filecontents*}{vilbert/exp4b_google.csv}
test_name,esize,pval
weat3:mask_v_word,0.3109862662106376,0.00002
weat3:mask_v_sent,0.3090947515676539,0.00001
weat6:mask_v_word,0.9578809684288061,0.00001
weat6:mask_v_sent,0.8698683597005572,0.00001
weat8:mask_v_word,0.7022135566134602,0.00003
weat8:mask_v_sent,0.3874083422021554,0.00001
weat11:mask_v_word,-1.071204721571895,1.0
weat11:mask_v_sent,-1.1132496785845496,1.0
weat12:mask_v_word,-0.00960304738771991,0.54741
weat12:mask_v_sent,0.07915949124679726,0.10439
weat13:mask_v_word,0.5181726362007267,0.00001
weat13:mask_v_sent,0.3315381150552869,0.00001
heilman_MF_double_bind_competent:mask_v_word,-0.5354324873013976,0.99901
heilman_MF_double_bind_competent:mask_v_sent,-0.3476805304252887,0.97433
heilman_MF_double_bind_likable:mask_v_word,-1.3439399952375184,1.0
heilman_MF_double_bind_likable:mask_v_sent,-0.11420619646210386,0.73924
heilman_race_double_bind_competent:mask_v_word,1.2372419683874651,0.00001
heilman_race_double_bind_competent:mask_v_sent,1.1499376982143668,0.00001
heilman_race_double_bind_likable:mask_v_word,0.5839681943843668,0.00038
heilman_race_double_bind_likable:mask_v_sent,0.7272060832219551,0.00002
angry_race:mask_v_word,0.418319135074014,0.05248
angry_race:mask_v_sent,0.4297382291101202,0.00001
occ_gender:mask_v_word,0.8982156138710684,0.00001
occ_gender:mask_v_sent,0.9056831221724555,0.00001
occ_race:mask_v_word,-0.3640902494118137,0.99992
occ_race:mask_v_sent,-0.29784500024170324,0.95373
\end{filecontents*}

\begin{filecontents*}{vilbert/exp1_concap.csv}
test_name,esize,pval
occ_gender:word,0.7485696952237365,0.00003
occ_gender:sent,0.7285556356419551,0.00001
occ_gender:contextual,0.5824360386799325,0.00001
weat6:word,0.942173114811451,0.00001
weat6:sent,1.1066053116793935,0.00001
weat6:contextual,0.7954808789284689,0.00001
\end{filecontents*}

\begin{filecontents*}{vilbert/exp2_concap.csv}
test_name,esize,pval
occ_gender:word,-0.5215333304552003,0.99845
occ_gender:sent,-0.30487076056057544,0.99312
occ_gender:contextual,1.9629746513754538,0.00001
weat6:word,0.9471492151346181,0.00001
weat6:sent,0.8310948076837256,0.00001
weat6:contextual,0.5824825096393175,0.00032
\end{filecontents*}

\begin{filecontents*}{vilbert/exp4a_concap.csv}
test_name,esize,pval
occ_gender:mask_t_word,0.7331815988177839,0.00001
occ_gender:mask_t_sent,0.7352537422606292,0.00001
weat6:mask_t_word,0.9550598090259129,0.00001
weat6:mask_t_sent,1.1487781974792837,0.00001
\end{filecontents*}

\begin{filecontents*}{vilbert/exp4b_concap.csv}
test_name,esize,pval
occ_gender:mask_v_word,0.7036867168576165,0.00002
occ_gender:mask_v_sent,0.7131722895626166,0.00001
weat6:mask_v_word,0.8480993309091281,0.00001
weat6:mask_v_sent,1.0893662902878734,0.00001
\end{filecontents*}

\begin{filecontents*}{lxmert/exp1_google.csv}
test_name,esize,pval
weat3:word,-0.1569,0.9764
weat3:sent,0.1906,0.00108
weat3:contextual,0.4386,0.00001
weat6:word,0.5544,0.00018
weat6:sent,0.6940,0.00001
weat6:contextual,0.1650,0.00165
weat8:word,0.4342357814311981,0.00498
weat8:sentence,0.03575807437300682,0.29061
weat8:contextual,0.1313689798116684,0.02126
weat11:word,-0.0787,0.6892
weat11:sent,-0.2019,0.99989
weat11:contextual,0.2501,0.00002
weat12:word,-0.0442,0.71318
weat12:sent,-0.3213,1.0
weat12:contextual,0.8773,0.00001
weat13:word,0.5806,0.00001
weat13:sent,0.1642,0.00419
weat13:contextual,0.9175,0.00001
heilman_MF_double_bind_competent:sent,-0.5496,0.99981
heilman_MF_double_bind_competent:contextual,-0.4848,0.99918
heilman_MF_double_bind_likable:sent,0.5997,0.00003
heilman_MF_double_bind_likable:contextual,-0.8324,1.0
heilman_MF_double_bind_competent:word,-1.1759,1.0
heilman_MF_double_bind_likable:word,-1.1032,1.0
heilman_race_double_bind_competent:word,0.9785,0.00001
heilman_race_double_bind_competent:sent,1.3005,0.00001
heilman_race_double_bind_competent:contextual,1.4577,0.00001
heilman_race_double_bind_likable:word,0.9301,0.00001
heilman_race_double_bind_likable:sent,-0.0410,0.60083
heilman_race_double_bind_likable:contextual,1.3950,0.00001
occ_gender:word,1.5630,0.00001
occ_gender:sent,1.3229,0.00001
occ_gender:contextual,0.5181,0.00152
angry_race:word,-1.3134,1.0
angry_race:sent,-0.1201,0.90438
angry_race:contextual,1.2740,0.00001
occ_race:word,-0.7107,1.0
occ_race:sent,-0.40126967430114746,0.98872
occ_race:contextual,-1.1087301969528198,1.0
\end{filecontents*}

\begin{filecontents*}{lxmert/exp2_google.csv}
test_name,esize,pval
weat3:word,0.6003,0.00001
weat3:sent,0.8446,0.00001
weat3:contextual,0.5789,0.00001
weat6:word,-0.1961,0.89232
weat6:sent,0.7764,0.00001
weat6:contextual,0.2083,0.00007
weat8:word,0.4342363774776459,0.00477
weat8:sentence,0.03575809672474861,0.28918
weat8:contextual,0.13136889040470123,0.02037
weat11:word,-0.1277,0.78883
weat11:sent,-0.5484,1.0
weat11:contextual,0.3811,0.00001
weat12:word,0.9343,0.00001
weat12:sent,0.3274,0.00001
weat12:contextual,0.9768,0.00001
weat13:word,-0.3750,1.0
weat13:sent,-0.1297,0.98156
weat13:contextual,0.9670,0.00001
heilman_MF_double_bind_competent:sent,-0.9781,1.0
heilman_MF_double_bind_competent:contextual,-1.1066,1.0
heilman_MF_double_bind_likable:sent,-0.8098,1.0
heilman_MF_double_bind_likable:contextual,-1.0017,1.0
heilman_MF_double_bind_competent:word,-1.3695,1.0
heilman_MF_double_bind_likable:word,-0.9250,1.0
heilman_race_double_bind_competent:word,1.0611,0.00001
heilman_race_double_bind_competent:sent,1.2481,0.00001
heilman_race_double_bind_competent:contextual,1.4592,0.00001
heilman_race_double_bind_likable:word,1.2869,0.00001
heilman_race_double_bind_likable:sent,0.4251,0.00346
heilman_race_double_bind_likable:contextual,1.3964,0.00001
occ_gender:word,1.3317,0.00001
occ_gender:sent,0.5774,0.00048
occ_gender:contextual,0.3402,0.02689
angry_race:word,-0.2659,0.84663
angry_race:sent,0.3053,0.00033
angry_race:contextual,1.3369,0.00001
occ_race:word,-0.3252,0.9999
occ_race:sent,0.2185947746038437,0.10837
occ_race:contextual,-1.0993893146514893,1.0
\end{filecontents*}

\begin{filecontents*}{lxmert/exp4a_google.csv}
test_name,esize,pval
weat3:mask_t_word,0.8179,0.00001
weat3:mask_v_word,-0.1197,0.93585
weat3:mask_t_sent,0.3283,0.00001
weat3:mask_v_sent,0.2067,0.00037
weat6:mask_t_word,1.3269,0.00001
weat6:mask_v_word,0.5673,0.00011
weat6:mask_t_sent,1.1789,0.00001
weat6:mask_v_sent,0.6851,0.00001
weat8:mask_t_word,0.5091,0.00506
weat8:mask_v_word,0.7349,0.00007
weat8:mask_t_sent,0.10729531198740005,0.04713
weat8:mask_v_sent,0.03746335953474045,0.28319
weat11:mask_t_word,-1.3309,1.0
weat11:mask_v_word,-0.1234,0.78119
weat11:mask_t_sent,-1.3278,1.0
weat11:mask_v_sent,-0.2249,0.99996
weat12:mask_t_word,-0.4221,1.0
weat12:mask_v_word,-0.0852,0.85949
weat12:mask_t_sent,-0.3864,1.0
weat12:mask_v_sent,-0.3599,1.0
weat13:mask_t_word,0.2065,0.00435
weat13:mask_v_word,0.6030,0.00001
weat13:mask_t_sent,-0.1040,0.95033
weat13:mask_v_sent,0.1717,0.00305
heilman_MF_double_bind_competent:mask_t_sent,-0.2134,0.91009
heilman_MF_double_bind_competent:mask_v_sent,-0.5513,0.99985
heilman_MF_double_bind_likable:mask_t_sent,0.2815,0.03805
heilman_MF_double_bind_likable:mask_v_sent,0.7186,0.00001
heilman_MF_double_bind_competent:mask_t_word,-0.0349,0.58851
heilman_MF_double_bind_competent:mask_v_word,-1.1706,1.0
heilman_MF_double_bind_likable:mask_t_word,-1.1029,1.0
heilman_MF_double_bind_likable:mask_v_word,-1.1138,1.0
heilman_race_double_bind_competent:mask_t_word,1.3060,0.00001
heilman_race_double_bind_competent:mask_v_word,0.9730,0.00001
heilman_race_double_bind_competent:mask_t_sent,1.3316,0.00001
heilman_race_double_bind_competent:mask_v_sent,1.2935,0.00001
heilman_race_double_bind_likable:mask_t_word,1.0570,0.00001
heilman_race_double_bind_likable:mask_v_word,0.9269,0.00001
heilman_race_double_bind_likable:mask_t_sent,0.1144,0.23833
heilman_race_double_bind_likable:mask_v_sent,-0.1374,0.80488
occ_gender:mask_t_word,0.7983,0.00001
occ_gender:mask_v_word,1.5758,0.00001
occ_gender:mask_t_sent,0.9219,0.00001
occ_gender:mask_v_sent,1.3225,0.00001
angry_race:mask_t_word,0.9199,0.00012
angry_race:mask_v_word,-1.2475,1.0
angry_race:mask_t_sent,-0.0093,0.54386
angry_race:mask_v_sent,-0.1309,0.92409
occ_race:mask_t_word,-1.0055,1.0
occ_race:mask_v_word,-0.7150,1.0
occ_race:mask_t_sent,-0.6451810598373413,0.99987
occ_race:mask_v_sent,-0.3803696036338806,0.98521
\end{filecontents*}

\begin{filecontents*}{lxmert/exp4b_google.csv}
test_name,esize,pval
\end{filecontents*}

\begin{filecontents*}{lxmert/exp1_coco.csv}
test_name,esize,pval
occ_gender:word,0.3892,0.00643
occ_gender:sent,-0.1786,0.92316
occ_gender:contextual,-0.1368,0.85979
weat6:word,0.9177,0.00001
weat6:sent,1.3192,0.00001
weat6:contextual,1.5274,0.00001
\end{filecontents*}

\begin{filecontents*}{lxmert/exp2_coco.csv}
test_name,esize,pval
occ_gender:word,-0.6564,1.0
occ_gender:sent,-1.1370,1.0
occ_gender:contextual,-0.7040,1.0
weat6:word,0.6135,0.00003
weat6:sent,1.1601,0.00001
weat6:contextual,1.4588,0.00001
\end{filecontents*}

\begin{filecontents*}{lxmert/exp4a_coco.csv}
test_name,esize,pval
occ_gender:mask_t_word,-0.1866,0.87985
occ_gender:mask_v_word,0.3473,0.01371
occ_gender:mask_t_sent,-0.0748,0.72676
occ_gender:mask_v_sent,-0.1713,0.91333
weat6:mask_t_word,-0.1001,0.73576
weat6:mask_v_word,0.8950,0.00001
weat6:mask_t_sent,0.0065,0.48163
weat6:mask_v_sent,1.3200,0.00001
\end{filecontents*}

\begin{filecontents*}{lxmert/exp4b_coco.csv}
test_name,esize,pval
\end{filecontents*}

\begin{filecontents*}{vlbert/exp1_google.csv}
test_name,esize,pval
weat3:word,1.3710,0.00001
weat3:sent,0.9251017135911772,0.00001
weat3:contextual,0.6767551637395609,0.00001
weat6:word,1.607107972783644,0.00001
weat6:sent,-0.01768657146974929,0.62297
weat6:contextual,0.4596425713357045,0.00001
weat8:word,-0.29439665981863933,0.9584
weat8:sent,0.18829469167535245,0.00155
weat8:contextual,0.2612965078717853,0.00002
weat11:word,-1.1951557956598695,1.0
weat11:sent,0.014926862465554802,0.3969
weat11:contextual,-0.18304161216714807,0.99938
weat12:word,-1.4450799446000067,1.0
weat12:sent,-0.9575629263570352,1.0
weat12:contextual,0.08291804606162173,0.09082
weat13:word,-1.4447647641849823,1.0
weat13:sent,0.9838503105829991,0.00001
weat13:contextual,0.8984091138725172,0.00001
heilman_MF_double_bind_competent:word,-1.2797753810882568,1.0
heilman_MF_double_bind_competent:sent,-1.3452966630172944,1.0
heilman_MF_double_bind_competent:contextual,0.3086803096626917,0.02481
heilman_MF_double_bind_likable:word,-0.9127296805381775,1.0
heilman_MF_double_bind_likable:sent,-0.029003173750248117,0.57238
heilman_MF_double_bind_likable:contextual,-0.1870383090192114,0.88122
heilman_race_double_bind_competent:word,1.4356,0.00001
heilman_race_double_bind_competent:sent,1.4776,0.00001
heilman_race_double_bind_competent:contextual,1.5364,0.00001
heilman_race_double_bind_likable:word,0.8720,0.00001
heilman_race_double_bind_likable:sent,1.0081,0.00001
heilman_race_double_bind_likable:contextual,0.1223,0.22132
occ_gender:word,0.9961,0.00001
occ_gender:sent,-0.0044,0.58873
occ_gender:contextual,0.1076,0.2738
angry_race:word,1.5891,0.00001
angry_race:sent,-0.4812,1.0
angry_race:contextual,-0.1289,0.91985
occ_race:word,1.3834,1.0
occ_race:sent,-0.05784434452652931,0.62968
occ_race:contextual,0.11565955728292465,0.25923
\end{filecontents*}

\begin{filecontents*}{vlbert/exp2_google.csv}
test_name,esize,pval
weat3:word,1.3411,0.00001
weat3:sent,-0.08006883292159772,0.90074
weat3:contextual,0.7643554769344945,0.00001
weat6:word,1.9714351005203867,0.00001
weat6:sent,1.5704188084341262,0.00001
weat6:contextual,0.44427853779951637,0.00001
weat8:word,-0.29439665981863933,0.96001
weat8:sent,0.18829469167535245,0.00188
weat8:contextual,0.2612965078717853,0.00004
weat11:word,-0.7665859320524779,0.99999
weat11:sent,-0.2139087307514898,0.99995
weat11:contextual,-0.17176000960568472,0.99898
weat12:word,-1.4879494460965152,1.0
weat12:sent,-1.271881990482511,1.0
weat12:contextual,0.1784313595836379,0.00198
weat13:word,-1.5088498353391555,1.0
weat13:sent,0.9513681671821248,0.00001
weat13:contextual,0.9600825446916718,0.00001
heilman_MF_double_bind_competent:word,1.5035549402236938,0.00001
heilman_MF_double_bind_competent:sent,-1.1353920412463634,1.0
heilman_MF_double_bind_competent:contextual,0.4371938963327384,0.00266
heilman_MF_double_bind_likable:word,-1.9844017028808594,1.0
heilman_MF_double_bind_likable:sent,1.9906884134287015,0.00001
heilman_MF_double_bind_likable:contextual,-0.11797466337493868,0.77048
heilman_race_double_bind_competent:word,1.4056,0.00001
heilman_race_double_bind_competent:sent,1.4529,0.00001
heilman_race_double_bind_competent:contextual,1.5665,0.00001
heilman_race_double_bind_likable:word,0.8978,0.00001
heilman_race_double_bind_likable:sent,1.2944,0.00001
heilman_race_double_bind_likable:contextual,0.0588,0.35384
occ_gender:word,-1.7442,1.0
occ_gender:sent,-0.2012,0.8803
occ_gender:contextual,0.1619,0.18408
angry_race:word,1.6730,0.00001
angry_race:sent,0.0286,0.37733
angry_race:contextual,-0.2069,0.98835
occ_race:word,-1.3951,1.0
occ_race:sent,-0.028394222259521484,0.56466
occ_race:contextual,0.10391989350318909,0.28142
\end{filecontents*}

\begin{filecontents*}{vlbert/exp4a_google.csv}
test_name,esize,pval
weat3:mask_t_sent,-0.007321442011743784,0.54663
weat3:mask_v_sent,0.947357714176178,0.00001
weat6:mask_t_sent,-0.002257743151858449,0.51526
weat6:mask_v_sent,-0.030683182179927826,0.70949
weat8:mask_t_sent,0.26605549454689026,0.00004
weat8:mask_v_sent,0.1797257512807846,0.00271
weat11:mask_t_sent,0.030167343094944954,0.29385
weat11:mask_v_sent,0.01850028522312641,0.3731
weat12:mask_t_sent,0.004776771180331707,0.46893
weat12:mask_v_sent,-1.0624783039093018,1.0
weat13:mask_t_sent,-0.0002684929932001978,0.50203
weat13:mask_v_sent,0.9455083608627319,0.00001
heilman_MF_double_bind_competent:mask_t_sent,-1.9937399625778198,1.0
heilman_MF_double_bind_competent:mask_v_sent,-1.0484533309936523,1.0
heilman_MF_double_bind_likable:mask_t_sent,-1.993740200996399,1.0
heilman_MF_double_bind_likable:mask_v_sent,0.6444356441497803,0.00002
heilman_race_double_bind_competent:mask_t_sent,-1.993740439414978,1.0
heilman_race_double_bind_competent:mask_v_sent,1.449755072593689,0.00001
heilman_race_double_bind_likable:mask_t_sent,-1.9937405586242676,1.0
heilman_race_double_bind_likable:mask_v_sent,1.0640593767166138,0.00001
occ_gender:mask_t_sent,-0.17357811331748962,0.83496
occ_gender:mask_v_sent,0.002606247318908572,0.47976
occ_race:mask_t_sent,-0.17179498076438904,0.83443
occ_race:mask_v_sent,-0.2547789514064789,0.92372
angry_race:mask_t_sent,-0.42348161339759827,1.0
angry_race:mask_v_sent,-0.07773447781801224,0.80345
\end{filecontents*}

\begin{filecontents*}{vlbert/exp4b_google.csv}
test_name,esize,pval
\end{filecontents*}

\begin{filecontents*}{vlbert/exp1_concap.csv}
test_name,esize,pval
occ_gender:word,-0.3146,0.97717
occ_gender:sent,-0.0097,0.53146
occ_gender:contextual,0.0136,0.45731
weat6:word,-0.1408,0.81345
weat6:sent,0.0021,0.47475
weat6:contextual,0.6081,0.00016
\end{filecontents*}

\begin{filecontents*}{vlbert/exp2_concap.csv}
test_name,esize,pval
occ_gender:word,1.9937,0.00001
occ_gender:sent,0.6888,0.00001
occ_gender:contextual,0.8998,0.00001
weat6:word,1.9759,0.00001
weat6:sent,-1.1661,1.0
weat6:contextual,0.0007,0.49536
\end{filecontents*}

\begin{filecontents*}{vlbert/exp4a_concap.csv}
test_name,esize,pval
occ_gender:mask_t_sent,0,0.49721
occ_gender:mask_v_sent,-0.0033,0.51683
weat6:mask_t_sent,0,0.49866
weat6:mask_v_sent,-0.0035,0.51645
\end{filecontents*}

\begin{filecontents*}{vlbert/exp4b_concap.csv}
test_name,esize,pval
\end{filecontents*}

\maketitle
\begin{abstract}
We generalize the notion of measuring social biases in word embeddings to visually grounded word embeddings.
  Biases are present in grounded embeddings, and indeed seem to be equally or
  more significant than for ungrounded embeddings.
  This is despite the fact that vision and language can suffer from different
  biases, which one might hope could attenuate the biases in both.
  Multiple ways exist to generalize metrics measuring bias in word embeddings to this new setting.
  We introduce the space of generalizations (Grounded-WEAT and Grounded-SEAT)
  and demonstrate that three generalizations answer different yet important
  questions about how biases, language, and vision interact.
  These metrics are used on a new dataset, the first for grounded bias, created
  by augmenting standard linguistic bias benchmarks with 10,228 images from
  COCO, Conceptual Captions, and Google Images.
  Dataset construction is challenging because vision datasets are themselves very biased.
  The presence of these biases in systems will begin to have real-world
  consequences as they are deployed, making carefully measuring bias and then
  mitigating it critical to building a fair society.
\end{abstract}

\section{Introduction}
Since the introduction of the Implicit Association Test (IAT) by
\citet{greenwald1998measuring}, we have had the ability to measure biases in
humans.
Many IAT tests focus on social biases, such as inherent beliefs about someone based on their racial or gender identity.
Social biases have negative implications for the most marginalized people, \eg applicants perceived to be Black based on their names are less likely to receive job interview callbacks than their white counterparts \cite{bertrand2004emily}.

\citet{caliskan2017semantics} introduce an
equivalent of the IAT for word embeddings, 
called the Word Embedding Association Test (WEAT), to measure word associations between concepts.
%
The results of testing bias in word embeddings using WEAT parallel those seen
when testing humans: both reveal many of the same biases with similar
significance.
\citet{may2019measuringbias} extend this work with a metric called the Sentence
Encoder Association Test (SEAT), that probes biases in embeddings of sentences instead
of just words.
We take the next step and demonstrate how to test visually grounded embeddings,
specifically embeddings from visually-grounded BERT-based models by extending prior work into what we term
Grounded-WEAT and Grounded-SEAT.
The models we evaluate are ViLBERT~\citep{lu2019vilbert},
VisualBERT~\citep{li2019visualbert} , LXMert~\citep{tan2019lxmert} and VL-BERT~\citep{su2019vl}.

Grounded embeddings are used for many consequential tasks in natural language
processing, like visual dialog~\citep{murahari2019large} and visual question
answering~\citep{hu2019iterative}.
Many real-world tasks such as scanning documents and interpreting images in
context employ joint embeddings as the performance gains are significant over
using separate embeddings for each modality.
It is therefore important to measure the biases of these grounded embeddings.
Specifically, we seek to answer three questions:

\emph{Do joint embeddings encode social biases?}
Since visual biases can be different from those in language, we would expect to see
a difference in the biases exhibited by grounded embeddings.
%
Biases in one modality might dampen or amplify the other.
%
We find equal or larger biases for grounded embeddings compared to the
ungrounded embeddings reported in \citet{may2019measuringbias}.
We hypothesize that this may be because visual datasets used to train multimodal models are much smaller and
much less diverse than language datasets.

\emph{Can grounded evidence that counters a stereotype alleviate biases?}
The advantage to having multiple modalities is that one modality can demonstrate
that a learned bias is irrelevant to the particular task being carried out.
For example, one might provide an image of a woman who is a doctor alongside a sentence about a doctor, and then
measure the bias against women doctors in the embeddings.
We find that the bias is largely not impacted, \ie direct visual evidence
against a bias helps little.

\begin{table*}[h!]
  \centering
  \small
  \begin{tabular}{cr|cr|cr}
    C3: EA/AA, (Un)Pleasant &1648&C6: M/W, Career/Family&780&C8: Science/Arts, M/W&718\\
    C11: M/W, (Un)Pleasant&1680&+C12: EA/AA, Career/Family&748&+C13: EA/AA, Science/Arts&522\\
    DB: M/W, Competent&560&DB: M/W, Likeable&480&M/W, Occupation&960\\
    +DB: EA/AA, Competent&440&+DB: EA/AA, Likeable&360&EA/AA, Occupation&928\\
    &&Angry Black Woman (ABW) &760&&\\
  \end{tabular}\\[0.5ex]
  (a) Number of images for all bias tests in the dataset collected from Google Images.\\[1ex]
  \begin{tabular}{cr|cr}
    C6: M/W, Career/Family&254&M/W, Occupation&229
  \end{tabular}\\[0.5ex]
  (b) Number of images for bias tests in the dataset collected from COCO.\\[1ex]
  \begin{tabular}{cr|cr}
    C6: M/W, Career/Family&203&M/W, Occupation&171
  \end{tabular}\\[0.5ex]
  (c) Number of images for bias tests in the dataset collected from Conceptual Captions.
  \caption{The number of images per bias test in our dataset (EA/AA=European
    American/African American names; M/W=names of men/women, renamed from M/F to reflect gender rather than sex). Tests prefixed by
    ``C'' are from \protect\cite{caliskan2017semantics}; \textit{Angry Black
      Woman (ABW)} and ``DB'' prefixes are from \protect\cite{may2019measuringbias};
    prefixes ``+C'' and ``+DB'' are from \protect\cite{tan2019assessing}. Each
    class contains an equal number of images per target-attribute pair. The
    dataset sourced from Google Images is complete, shown in (a).  Datasets
    sourced from COCO and Conceptual Captions, shown in (b) and (c)
    respectively, contain a subset of the tests because the lack of gender and
    racial diversity in these datasets makes creating balanced data for grounded
    bias tests impractical.}
  \label{tab:dataset-statistics}
\end{table*}

\emph{To what degree are biases encoded in grounded word embeddings from language or vision?}
It may be that grounded word embeddings derive all of their biases from one modality,
such as language.
In this case, vision would be relevant to the embeddings, but would not impact
the measured bias.
We find that, in general, both modalities contribute to encoded bias, but some model architectures are more dominated by language.
Vision could have a more substantial impact on grounded word embeddings.

We generalize WEAT and SEAT to grounded embeddings to answer these questions.
Several generalizations are possible, three of which correspond to the questions
above, while the rest appear unintuitive or redundant.
We first extracted images from COCO~\citep{chen2015microsoft} and Conceptual Captions~\citep{sharma2018conceptual};
the images and English captions in these datasets lack diversity, making finding data for most existing bias
tests nearly impossible.
To address this, we created an additional dataset from Google Images that depicts
the targets and attributes required for all bias tests considered.
This work does not attempt to reduce bias in grounded models.
We believe that the first critical step to doing so, is having metrics and a dataset to understand grounded biases, which we introduce here.

The dataset introduced along with the metrics presented can serve as a
foundation for future work to eliminate biases in grounded word embeddings.
In addition, they can be used as a sanity check before deploying systems
to understand what kinds of biases are present.
The relationship between linguistic and visual biases in humans is unclear, as
the IAT has not been used in this way.

Our contributions are:
\begin{compactenum}
\item Grounded-WEAT and Grounded-SEAT answering three questions about biases in
  grounded embeddings,
\item a new dataset for testing biases in grounded systems,
\item demonstrating that grounded word embeddings have social biases,
\item showing that grounded evidence has little impact on social biases, and
\item showing that biases come from a mixture of language and vision.
\end{compactenum}

\section{Related Work}

Models that compute word embeddings are widespread
\citep{mikolov2013distributed,devlin2018bert,peters2018deep,radford2018GPT}.
Given their importance, measuring the presence of harmful social biases in such models is critical.
\citet{caliskan2017semantics} introduce the Word Embedding Association Test,
WEAT, based on the Implicit Association Test, IAT, to measure biases in word
embeddings.
WEAT measures social biases using multiple tests that pair target
concepts, \eg gender, with attributes, \eg careers and families.

\citet{may2019measuringbias} generalize WEAT to biases in sentence embeddings,
introducing the Sentence Encoder Association Test (SEAT).
\citet{tan2019assessing} generalize SEAT to contextualized word representations,
\eg the encoding of a word in context in the sentence; \cite{zhao2019gender} also evaluated gender bias in contexutal embeddings from ELMo.
These advances are incorporated into the grounded metrics developed here, by
measuring the bias of word embeddings, sentence embeddings, as well as
contextualized word embeddings.

\citet{blodgett2020survey} provide an in-depth analysis of NLP papers exploring bias in datasets and models and also highlight key areas for improvement in approaches.
We point the reader to this paper and aim to draw from key suggestions from this work throughout.
%

\section{The Grounded WEAT/SEAT Dataset}

\begin{figure*}[t!]
  \centering
  \scalebox{0.9}{\begin{tabular}{cccc}
     \includegraphics[width=0.19\textwidth]{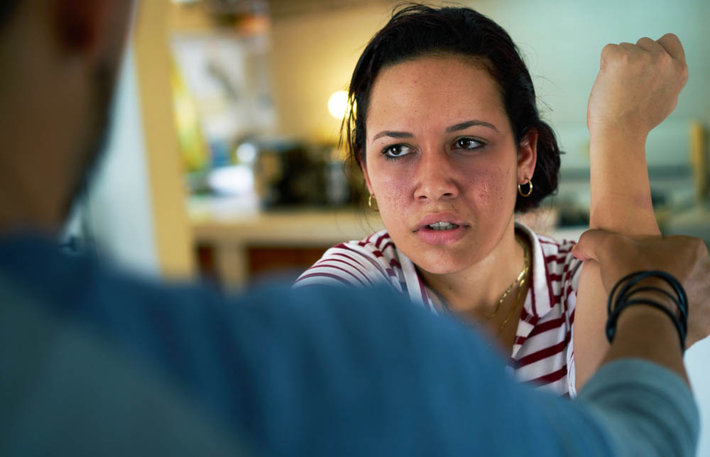}
    &\includegraphics[width=0.07\textwidth]{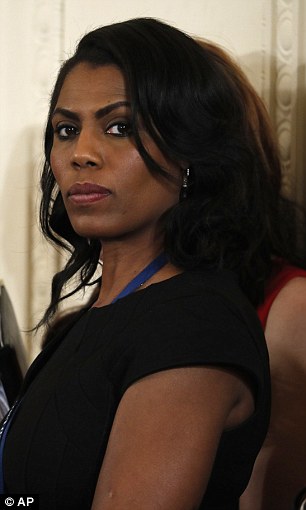}
    &\includegraphics[width=0.19\textwidth]{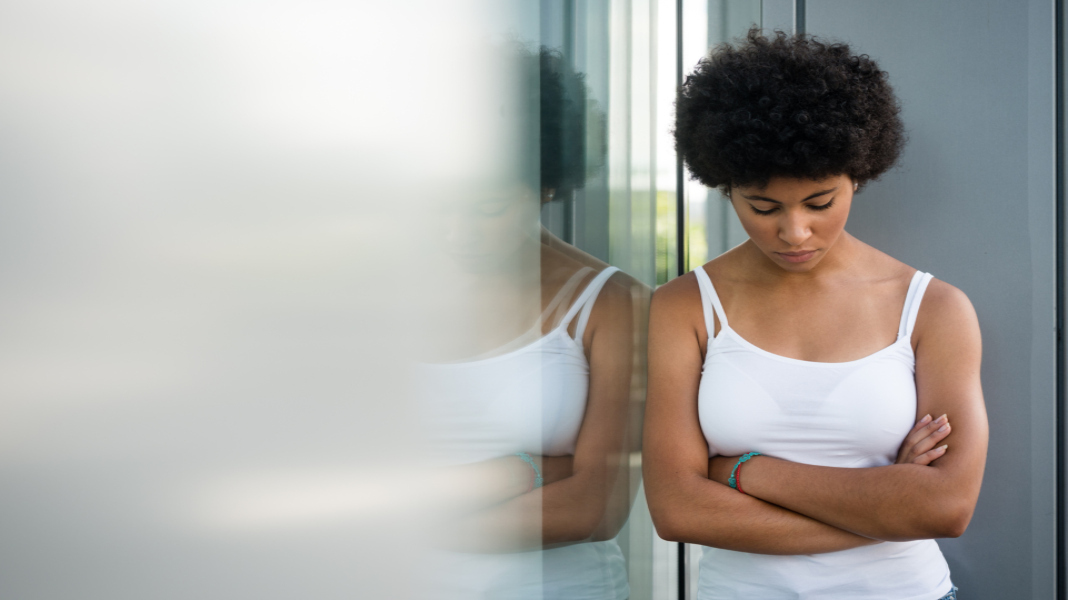}
    &\includegraphics[width=0.11\textwidth]{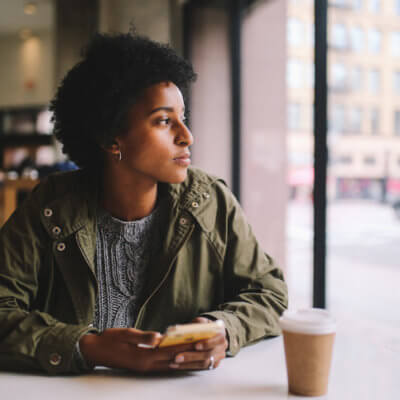}\\
    $a_x$
    &$a_x$
    &$b_x$
    &$b_x$\\
     \includegraphics[width=0.19\textwidth]{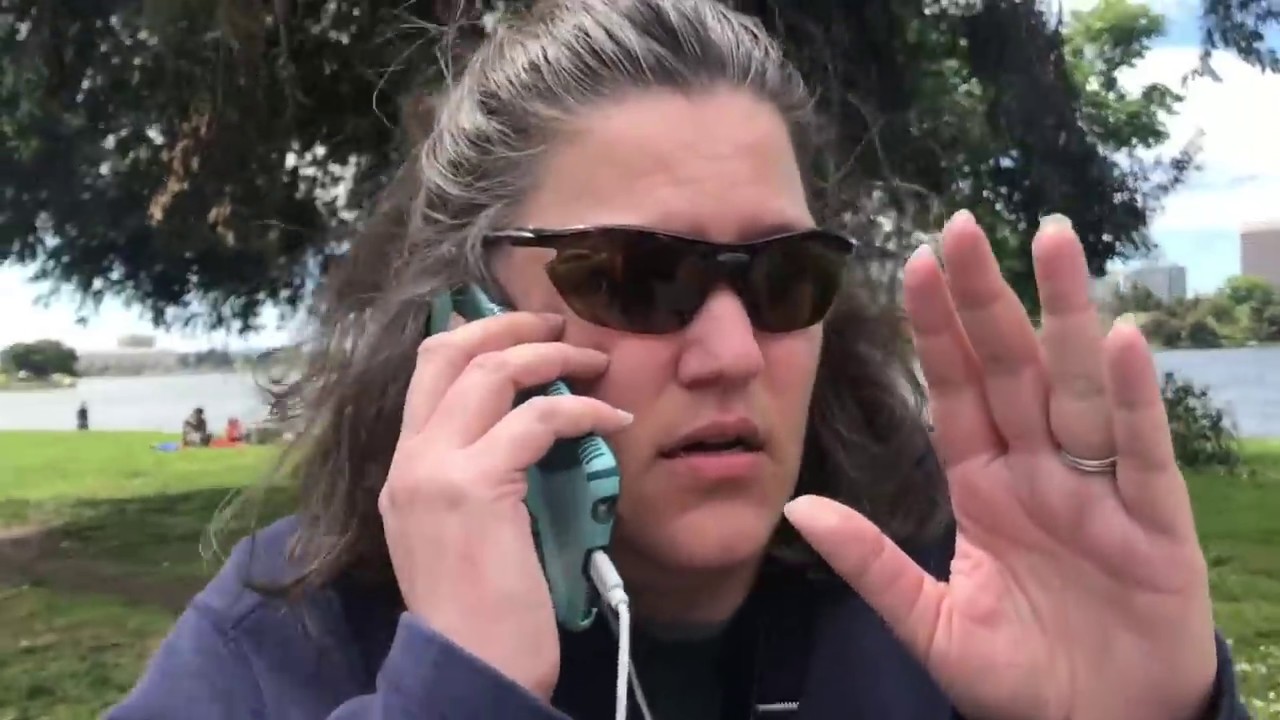}
    &\includegraphics[width=0.19\textwidth]{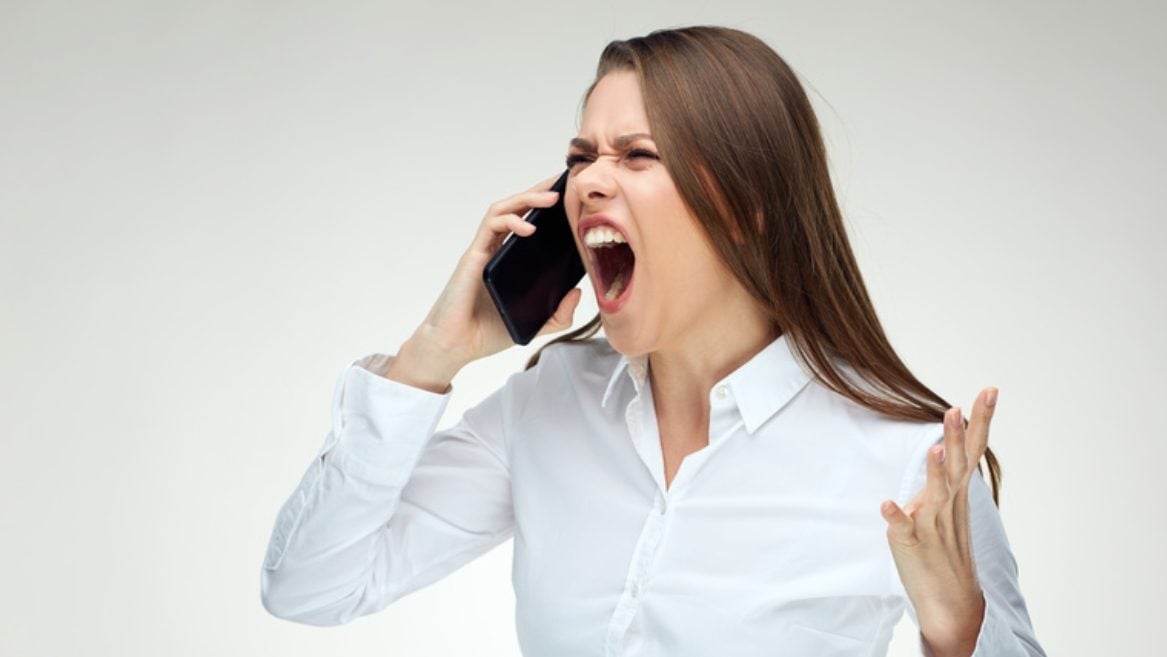}
    &\includegraphics[width=0.19\textwidth]{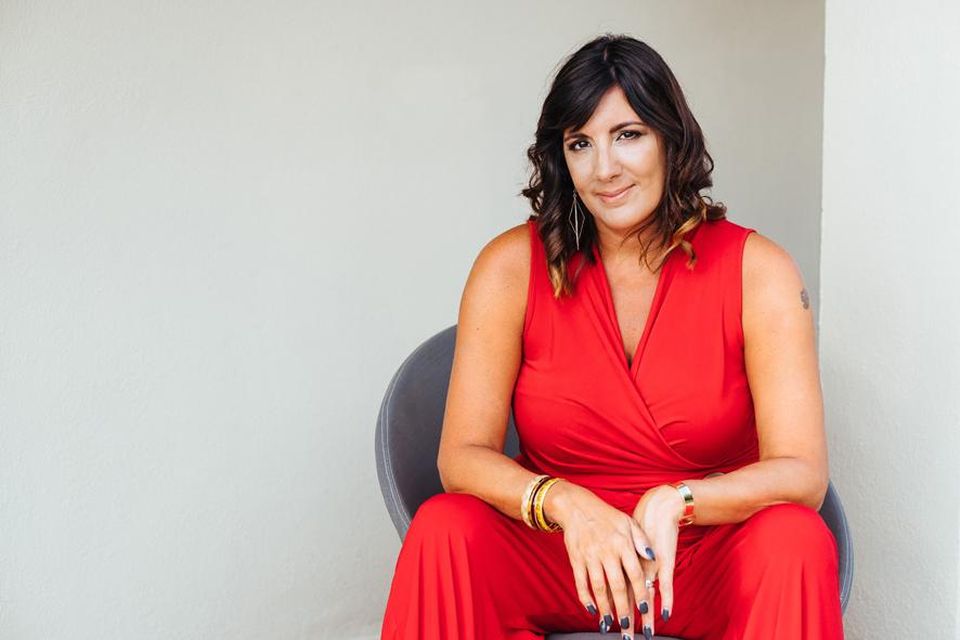}
    &\includegraphics[width=0.19\textwidth]{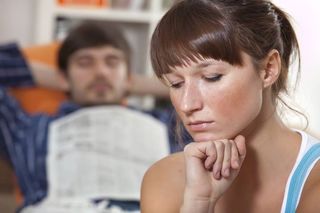}\\
    $a_y$
    &$a_y$
    &$b_y$
    &$b_y$\\
  \end{tabular}}
  \caption{One example set of images for the bias class \emph{Angry black women
      stereotype}~\citep{collins2004black}, where the targets, $X$ and $Y$, are
    typical names of \emph{black women} and \emph{white women}, and the linguistic attributes are
    \emph{angry} or \emph{relaxed}. The top row depicts black women; the bottom
    row depicts white women. The two left columns depict aggressive stances
    while the two right columns depict more passive stances. The attributes for
    the grounded experiment, $A_x$, $B_x$, $A_y$, and $B_y$, are images that
    depict a target and in the context of an attribute.}
  \label{fig:dataset-example}
\end{figure*}

\begin{table}[t]
  \small\centering
\begin{tabular}{crr}
  Embedding index & Word\\\hline
  1 & Man\\
  2 & Woman\\
  3 & Lawyer\\
  4 & Teacher\\
\end{tabular}\\
 (a) Possible embeddings for an ungrounded model\\[2ex]
\begin{tabular}{crr}
  Embedding index & Word   & What the image shows\\\hline
  1 & Man     & \emph{Any Man}\\
  2 & Man     & \emph{Any Woman}\\
  3 & Woman   & \emph{Any Man}\\
  4 & Woman   & \emph{Any Woman}\\
  5 & Lawyer  & \emph{Man Lawyer}\\
  6 & Lawyer  & \emph{Man Teacher}\\
  7 & Lawyer  & \emph{Woman Lawyer}\\
  8 & Lawyer  & \emph{Woman Teacher}\\
      9 & Teacher & \emph{Man Lawyer}\\
  10 & Teacher & \emph{Man Teacher}\\
  11 & Teacher & \emph{Woman Lawyer}\\
  12 & Teacher & \emph{Woman Teacher}\\
\end{tabular}\\
 (b) Possible embeddings for a visually grounded model
 \vspace{-1ex}
 \caption{The content of a trivial hypothetical grounded dataset to demonstrate
   the intuition behind the three experiments. The dataset could be used to
   answer questions about biases in association between gender and
   occupation. Each entry is an embedding that can be computed with an ungrounded model, (a),
   and with a grounded model, (b),
   for this hypothetical dataset. This demonstrates the additional degrees of
   freedom when evaluating bias in grounded datasets. In the subsections that
   correspond to each of the experiments, \cref{sec:exp1,sec:exp2,sec:exp3}, we
   explain which parts of this dataset are used in each
   experiment. Our experiments only use a subset of the possible embeddings,
   leaving room for new metrics that answer other questions.}
\label{tab:trivial-dataset}
\vspace{-1ex}
\end{table}

\nocite{zhao2018gender}

Existing WEAT/SEAT bias tests (\citet{caliskan2017semantics},
\citet{may2019measuringbias} and \citet{tan2019assessing}) contain sentences for categories and attributes; we augment these tests to a grounded domain by pairing each word/sentence with an image.
%
%
VisualBERT and ViLBERT were trained on COCO and Conceptual Captions respectively, so we use the images in these datasets' validation splits by querying the captions for the keywords.
%
To compensate for their lack of diversity, we collected another version of
the dataset where the images are top-ranked hits on Google Images.
Results on COCO and Conceptual Captions are still important for the bias tests that can be collected, for two reasons.
First, it gives us an indication of where datasets are lacking: the fact that images
cannot be sourced for so many tests means these datasets particularly lack representation for these identities.
Second, since COCO and Conceptual Captions form part of the training sets
for VisualBERT and ViLBERT, this ensures that biases are not
a property of poor out-of-domain generalization.
The differences in bias in-domain and out-of-domain appear to be small.
Images were collected prior to the implementation of the experiment.
We provide original links to all collected images and scripts to download them.
%

\section{Methods}
\label{sec:methods}

Existing WEAT/SEAT bias tests \citep{caliskan2017semantics} base the Word Embedding Association Test (WEAT) on
an IAT test administered to humans.
Two sets of target words, $X$ and $Y$, and two sets of attribute words, $A$ and $B$,
are used to probe systems.
The average cosine similarity between pairs of word embeddings is used as the
basis of an indicator of bias, as in:
\begin{equation}\label{eq:test-stat-text-single}
  s(w,A,B) = \mean_{a\in A} \textrm{cos}(w,a) - \mean_{b\in B} \textrm{cos}(w,b)
\end{equation}
where $s$ measures how close on average the embedding for word $w$ is compared
to the words in attribute set $A$ and attribute set $B$.
%
%
Such relative distances between word vectors indicate how related two concepts are and are directly used in many natural language processing
tasks, \eg analogy completion~\citep{drozd2016word}.

By incorporating both target word classes $X$ and $Y$, this distance can be used
to measure bias.
The space of embeddings may encode social biases by making some targets, \eg  men's
names or women's names, closer to one profession than another.
In this case, bias is defined as one of the two targets being significantly closer to one
set of socially stereotypical attribute words compared to the other.
%
%
The test in \cref{eq:test-stat-text-single} is computed for each set of targets,
determining their relative distance to the attributes.
The difference between the target distances reveals which target sets are more
associated with which attribute sets:
\begin{equation}\label{eq:test}
  s(X,Y,A,B) = \sum_{x\in X}s(x,A,B) - \sum_{y\in Y}s(y,A,B)
\end{equation}

The effect size, \ie the number of standard deviations in which the peaks of the
distributions of embedding distances differ, of this metric is computed as:
\begin{equation}\label{eq:ungrounded-effect-size}
  d= \frac{\displaystyle\mean_{x\in X} s(x,A,B) - \displaystyle\mean_{y\in Y}s(y,A,B)}
  {\displaystyle\stddev_{w\in X \cup Y} s(w,A,B)}
\end{equation}

\citet{may2019measuringbias} extend this test to measure sentence embeddings, by
using sentences in the target and attribute sets.
\citet{tan2019assessing} extend the test to measure contextual effects, by
extracting the embedding of single target and attribute tokens in the context of
a sentence rather than the encoding of the entire sentence.
We demonstrate how to extend these notions to a grounded setting, which
naturally adapts these two extensions to the data, but requires new metrics
because vision adds new degrees of freedom to what we can measure.

To explain the intuition behind why multiple grounded tests are possible,
consider a trivial hypothetical dataset that measures only a single property;
see \cref{tab:trivial-dataset}.
This dataset is complete: it contains the cross product of every target category, \ie gender, and attribute category, \ie occupation, that can
happen in its minimal world.
In the ungrounded setting, only 4 embeddings can be computed because the
attributes are independent of the target category.
In the grounded setting, by definition, the attributes are words and images that correspond to one of the target categories.
This leads to 12 possible grounded embeddings\footnote{An
  alternate way to construct such a dataset might have ambiguity about which
  of two agents a sentence is referring to, more closely mirroring how language
  is used. This would require images that simultaneously depict both targets,
  \eg both a man and woman who are teachers. Finding such data is difficult and may be impossible in many cases, but
  it would also be a less realistic measure of bias. In practice, systems built on top of grounded embeddings will not be used with balanced images, and so while in a sense more
  elegant, this construction may completely misstate the biases one would see in
  the real world.}; see \cref{tab:trivial-dataset}.
We subdivide the attributes $A$ and $B$ into two categories, $A_x$ and $B_x$,
which depict the attributes with the category of target $x$, and $A_y$ and
$B_y$, with the category of target $y$.
%
%
Example images for the bias test for the intersectional
racial and gender stereotype that black women are inherently angry, are shown in~\cref{fig:dataset-example}.
These images depict the target's category and attributes; they are the
equivalent of the attributes in the ungrounded experiments.

With these additional degrees of freedom, we can formulate many different grounded tests
in the spirit of \cref{eq:test}.
We find that three such tests, described next, have intuitive explanations and
measure different but complementary aspects of bias in grounded word embeddings.
These questions are relevant to both bias and to the quality of word embeddings.
For example, attempting to measure the impact of vision separately from language on grounded word embeddings can
indicate if there is an over-reliance on one modality over another.

We evaluate bias tests on embeddings produced by Transformer-based vision and
language models which take as input an image and a caption.
Models are used to produce three kinds of embeddings (of single-word captions, full
sentence captions, and word embeddings in the context of a sentence) that are
each tested for biases.
These embeddings correspond to the hidden states of the language output of each
model.
For single-stream models like VisualBERT and VL-BERT, these are the hidden
states corresponding to the language token inputs.
For two-stream models like ViLBERT and LXMERT, these are the outputs of the
language Transformer.
When computing word and sentence embeddings, we follow
\citet{may2019measuringbias} and take the hidden state corresponding to the
\texttt{[CLS]} token (shown in blue in~\cref{fig:embeddings}).
When computing contextual embeddings, we follow \citet{tan2019assessing} and
take the embedding in the sequence corresponding to the token for the relevant
contextual word, e.g., for the sentence ``The \textit{man} is there'', we take
the embedding for the token ``man'' (shown in green in ~\cref{fig:embeddings}).
Note there can be multiple contextual tokens when a contextual word is subword
tokenized; we take the sequence corresponding to the first token.
To mask the language, every contextual token in the input is set to
\texttt{[MASK]}.
To mask the image, every region of interest or bounding box with a person label
is masked.
Models which did not use bounding boxes during training could not be included in
image masking tests.

\begin{figure*}[t!]
  \centering
  \scalebox{0.7}{%
  \begin{tabular}{cl}
    VisualBERT & \texttt{{\blue[CLS]} TOK0 ... {\green TOK\_CONTEXTUAL} ... TOKN [SEP] [IMG] IMG0 ... IMGN} \\
    VL-BERT & \texttt{{\blue [CLS]} TOK0 ... {\green TOK\_CONTEXTUAL} ... TOKN [SEP] IMG0 IMG1 ... IMGN [END]} \\
    ViLBERT & \texttt{{\blue [CLS]} TOK0 ... {\green TOK\_CONTEXTUAL} ... TOKN [SEP]} \\
    LXMert & \texttt{{\blue [CLS]} TOK0 ... {\green TOK\_CONTEXTUAL} ... TOKN [SEP] [CROSS\_MODAL]}
  \end{tabular}}
  \caption{Each row shows the output sequence corresponding to a given model's output. For ViLBERT and LXMERT, we only show the output of the language Transformer. For word and sentence embeddings, we take the encoding corresponding to the \texttt{[CLS]} token; for contextual embeddings, we take the encpding corresponding to the word in context, \texttt{[TOK\_CONTEXTUAL]}.}
  \label{fig:embeddings}
\end{figure*}

\subsection{Experiment 1: Do joint embeddings encode social biases?}
\label{sec:exp1}

This experiment measures biases by integrating out vision and looking at the resulting
associations.
For example, regardless of what the visual input is, are men deemed more likely to be
in some professions compared to women?
Similarly to \cref{eq:test}, we compute the association
between target concepts and attributes, except that we include all of the
images:
\begin{equation*}\begin{multlined}
  \hspace*{-1ex}s(X,Y,A,B) = \sum_{x\in X}s(x,A_x\cup A_y,B_x\cup B_y)\\
   - \sum_{y\in Y}s(y,A_x\cup A_y,B_x\cup B_y)
\end{multlined}\end{equation*}

To be concrete, for the trivial hypothetical dataset in
\cref{tab:trivial-dataset}, this corresponds to
$S(1,\{5,7\},\{10,12\}) - S(4,\{5,7\},\{10,12\})$, which compares the bias
relative to \emph{man} and \emph{woman} against \emph{lawyer} or \emph{teacher}
across all target images.
If no bias is present, we would expect the effect size to be zero.
Our hope would be that the presence of vision at training time would help
alleviate biases even if at test time any images are possible.

\subsection{Experiment 2: Can grounded evidence that counters a stereotype alleviate biases?}
\label{sec:exp2}
An advantage of grounded embeddings is that we can readily show
scenarios that clearly counter social stereotypes.
For example, the model may have a strong prior that men are more likely to have
some professions, but are the embeddings different when the visual input
provided shows women in those professions?
Similarly to \cref{eq:ungrounded-effect-size}, we compute the association
between target concept and attributes, except that we include only images
that correspond to the target concept's category:
{\begin{equation*}\begin{multlined}
  \hspace*{-2ex}s(X,Y,A,B) =\sum_{x\in X}s(x,A_x,B_x)\\ - \sum_{y\in Y}s(y,A_y,B_y)
\end{multlined}\end{equation*}}

To be concrete, for the trivial hypothetical dataset in
\cref{tab:trivial-dataset}, this corresponds to
$S(1,\{5\},\{10\}) - S(4,\{7\},\{12\})$, which computes the bias of \emph{man}
and \emph{woman} against \emph{lawyer} and \emph{teacher} relative to only
images that actually depict lawyers and teachers who are men when comparing to target
\emph{man} and lawyers and teachers who are women when comparing to target
\emph{woman}.
If no bias was present, we would expect the effect size to be zero.
Our hope would be that even if biases exist, clear grounded evidence to the
contrary would overcome them.

\subsection{Experiment 3: To what degree are biases encoded in grounded word embeddings from language or vision?}
\label{sec:exp3}

Even if biases exist, one might wonder how much of the bias comes from language and how much comes from vision?
Perhaps all of the biases come from language and vision only plays a small auxiliary role, or vice versa.
We can probe this question in at least two ways.
First, one could use images that are both congruent and incongruent with the stereotype.
We would in that case check if the model changes its embeddings in response to the congruent or incongruent images.
Similarly to \cref{eq:ungrounded-effect-size}, in this case we compute the association
between target concepts and attributes, except that we compare cases when images
support stereotypes to cases where images counter stereotypes and do not depict the target
concept:
{\begin{equation*}\begin{multlined}
  \hspace*{-0ex}s(X,Y,A,B) =\\
  \frac{1}{2}(|\sum_{x\in X}s(x,A_x,B_x) - \sum_{x\in X}s(x,A_y,B_y)|\\
  \hspace{2.5ex}+ |\sum_{y\in Y}s(y,A_y,B_y) - \sum_{y\in Y}s(y,A_x,B_x)|)
\end{multlined}\end{equation*}}
To be concrete, for the trivial hypothetical dataset in
\cref{tab:trivial-dataset}, this corresponds to
$\frac{1}{2}(|S(1,\{5\},\{10\}) - S(1,\{7\},\{12\})|+|S(2,\{7\},\{12\}) -
S(2,\{5\},\{10\})|)$, which compares the bias relative to \emph{man} against
\emph{lawyer} or \emph{teacher} and \emph{woman} against \emph{lawyer} or
\emph{teacher} relative to images that are either evidence for these occupations
as men and women.
%
We take the absolute value of the two, since they may be biased in different
ways.
If no bias was present, we would expect the effect size to be zero.

An alternate way to probe this bias makes use of the same test as in Experiment 2 with the addition of masking by taking advantage of how these models are pretrained with masked language tokens and masked image regions.
VisualBERT only uses masked language modeling and never masks image regions during training; it therefore cannot be probed using this method.
For each test, we alternatively mask either language tokens or image regions relevant to that specific test and measure the encoded bias.
When masking image regions we mask regions that contain people.
For example, in test C3, we mask every name and every pleasant or unpleasant term while token masking and every person while image masking.
This ablates the potential bias in one modality, allowing us to probe the other.

\section{Results}

We evaluate each model on images from the dataset used for pretraining and our collected images from Google Image search.
Pretraining datasets are MS-COCO for VisualBERT~\citep{li2019visualbert} and LXMert~\citep{tan2019lxmert} and Conceptual Captions for ViLBERT~\citep{lu2019vilbert} and VL-BERT~\citep{su2019vl}\footnote{Some pretraining images for VL-BERT are from the Visual Genome.}.
Image features are computed in the same manner as in the original publications.
We compute $p$-values using the updated permutation test described
in~\citet{may2019measuringbias}.
In each case, we evaluate the task-agnostic, pretrained base model without task-specific fine tuning.
The effect of task-specific training on biases is an interesting open question
for future work.

\begin{table*}[!t]
  \centering\scalebox{0.75}{\showExperiment{exp1}}
  \vspace{-0.5ex}
  \caption{The results for all bias classes on Experiment 1 using Google Images
    that asks \emph{Do joint embeddings encode social biases?} Numbers represent
    effect sizes and $p$-values for the permutation test described in
    \cref{sec:methods}. They are highlighted in blue when $p$-values are below
    0.05. Each bias type and model are tested three times against (W) word
    embeddings, (S) sentence embeddings, and (C) contextualized word embeddings.
    The answer to the question clearly appears to be yes. All models are biased. Note that
    out of domain, biases appear to be amplified.}
  \label{tab:exp1}
  \vspace{-0.5ex}
\end{table*}

\begin{table*}[!t]
  \centering\scalebox{0.75}{\showExperiment{exp2}}
  \vspace{-0.5ex}
  \caption{The results for all bias classes on Experiment 2 using Google Images
    that asks \emph{Can joint embeddings be shown grounded evidence that a bias
      does not apply?}  Numbers represent effect sizes and $p$-values for the
    permutation test described in \cref{sec:methods}. They are highlighted in
    blue when $p$-values are below 0.05. Each bias type and model are tested three
    times against (W) word embeddings, (S) sentence embeddings, and (C)
    contextualized word embeddings. The answer to the question appears to be no,
    although fewer tests are statistically significant compared to
    \cref{tab:exp1} showing that visual evidence is helpful.}
  \label{tab:exp2}
  \vspace{-0ex}
\end{table*}

\begin{table*}
  \hspace*{0em}\centering\scalebox{1}{
      \includegraphics*[width=\textwidth,clip,trim=0em 50em 0em 2em]{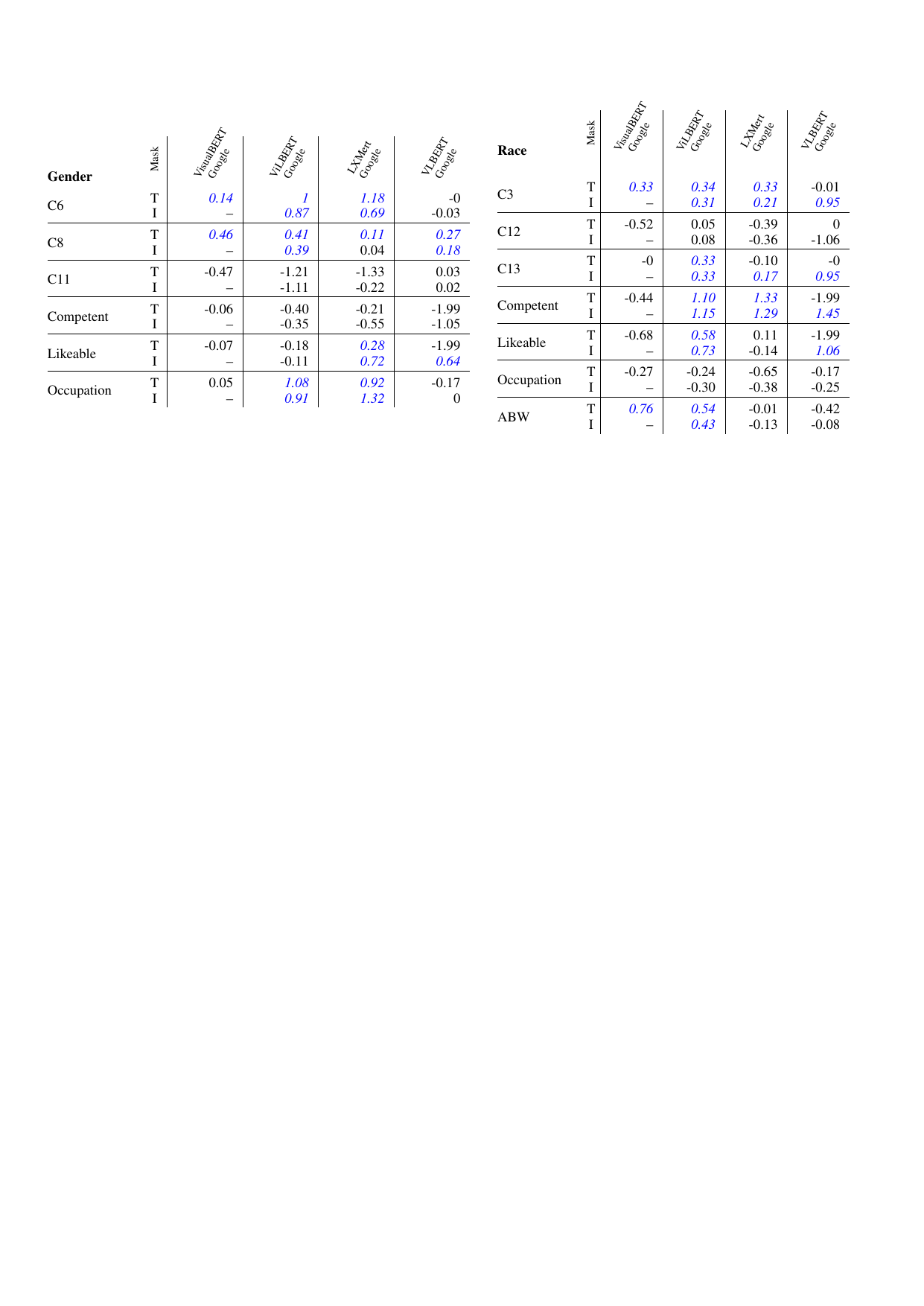}
    }\vspace{-1ex}
  \caption{The results for all bias classes on Experiment 3, using the second masking variant of the experiment, with Google Images
    asking the question \emph{To what degree are biases encoded in grounded word embeddings from language or vision?}  Numbers represent effect sizes and $p$-values for the
    permutation test described in \cref{sec:methods}. All numbers were measured over sentence-level encodings. They are highlighted in
    blue when $p$-values are below 0.05. Biases are measured for masked tokens (T) and masked image regions (I). This answer appears to be that both vision and language play a significant role, but this differs across model architectures.}
  \label{tab:exp3}
  \vspace{-1.5ex}
\end{table*}

\begin{table*}[t!]
  \hspace*{-6em}\centering\scalebox{1.2}{
      \includegraphics*[width=\textwidth,clip,trim=6em 50em 0em 10em]{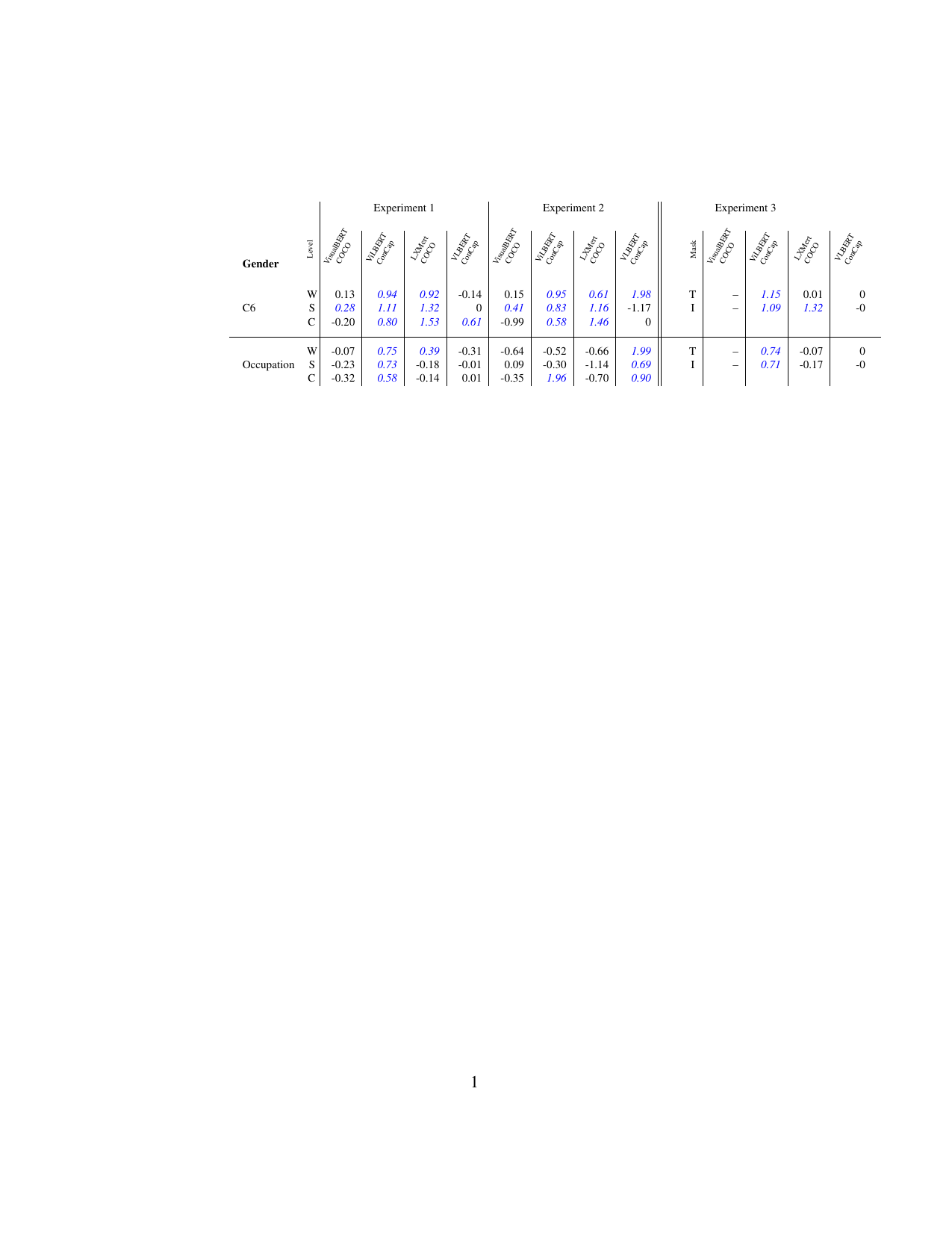}
    }\vspace{-1ex}
  \caption{The results for two classes of bias on all three experiments using
    COCO and Conceptual Captions. Images for other
    bias classes could not be found in these datasets. These results are
    generally consistent with results on the Google Images dataset.}
  \label{tab:expC}
  \vspace{-1.5ex}
\end{table*}

\begin{table*}[t!]

  \footnotesize
  \centering \textit{Number of statistically significant tests out of 6 total gender bias tests}\\
  \scalebox{0.8}{\hspace*{-4em}\begin{tabular*}{1\linewidth}{@{\extracolsep{\fill}}p{2em}p{0.5em}*{8}{|>{\raggedleft\arraybackslash}p{0.5em}}|*{4}{|>{\raggedleft\arraybackslash}p{0.5em}}|p{0.2em}}
  && \multicolumn{4}{c|}{Experiment 1}
  & \multicolumn{4}{c||}{Experiment 2}
  & \multicolumn{4}{c}{Experiment 3}\\[3ex]
  & \rotatebox{90}{\scalebox{0.8}{\parbox{4em}{\centering\footnotesize Level}}}
  & \hspace*{-3.8ex}\rotatebox{60}{\scalebox{0.8}{\parbox{4em}{\centering\footnotesize VisualBERT Google}}}
  & \hspace*{-3.8ex}\rotatebox{60}{\scalebox{0.8}{\parbox{4em}{\centering\footnotesize ViLBERT Google}}}
  & \hspace*{-3.8ex}\rotatebox{60}{\scalebox{0.8}{\parbox{4em}{\centering\footnotesize LXMert Google}}}
  & \hspace*{-3.8ex}\rotatebox{60}{\scalebox{0.8}{\parbox{4em}{\centering\footnotesize VLBERT Google}}}%
  
  & \hspace*{-3.8ex}\rotatebox{60}{\scalebox{0.8}{\parbox{4em}{\centering\footnotesize VisualBERT Google}}}
  & \hspace*{-3.8ex}\rotatebox{60}{\scalebox{0.8}{\parbox{4em}{\centering\footnotesize ViLBERT Google}}}%
  & \hspace*{-3.8ex}\rotatebox{60}{\scalebox{0.8}{\parbox{4em}{\centering\footnotesize LXMert Google}}}
  & \hspace*{-3.8ex}\rotatebox{60}{\scalebox{0.8}{\parbox{4em}{\centering\footnotesize VLBERT Google}}}%

  & \hspace*{-3.8ex}\rotatebox{60}{\scalebox{0.8}{\parbox{4em}{\centering\footnotesize Mask}}}
  & \hspace*{-3.8ex}\rotatebox{60}{\scalebox{0.8}{\parbox{4em}{\centering\footnotesize VisualBERT Google}}}
  & \hspace*{-3.8ex}\rotatebox{60}{\scalebox{0.8}{\parbox{4em}{\centering\footnotesize ViLBERT Google}}}
  & \hspace*{-3.8ex}\rotatebox{60}{\scalebox{0.8}{\parbox{4em}{\centering\footnotesize LXMert Google}}}%
  & \hspace*{-3ex}\rotatebox{60}{\scalebox{0.8}{\parbox{4em}{\centering\footnotesize VLBERT Google}}}
    \vspace{-3ex}\\%
  & W%
  & 2%
  & 2%
  & 3%
  & 3%
  & 2%
  & 2%
  & 2%
  & 2%
  &T%
  & -%
  & 1%
  & 3%
  & 4%
  \\%
  & {\leavevmode\hspace*{0.4ex}S}%
  & 2%
  & 1%
  & 3%
  & 3%
  & 3%
  & 3%
  & 2%
  & 2%
  &I%
  & -%
  & 2%
  & 3%
  & 3%
  \\%
  & {\leavevmode\hspace*{0.2ex}C}%
  & 3%
  & 3%
  & 3%
  & 4%
  & 2%
  & 3%
  & 3%
  & 4%
  &  %
  & %
  & 
  &
\\%
\end{tabular*}}%

\centering \vspace*{3ex}\textit{Number of statistically significant tests out of 7 total race bias tests}\\
\scalebox{0.8}{\hspace*{-4em}\begin{tabular*}{1\linewidth}{@{\extracolsep{\fill}}p{2em}p{0.5em}*{8}{|>{\raggedleft\arraybackslash}p{0.5em}}|*{4}{|>{\raggedleft\arraybackslash}p{0.5em}}|p{0.2em}}
  && \multicolumn{4}{c|}{Experiment 1}
  & \multicolumn{4}{c||}{Experiment 2}
  & \multicolumn{4}{c}{Experiment 3}\\[3ex]
  
  & \rotatebox{90}{\scalebox{0.7}{\parbox{4em}{\centering\footnotesize Level}}}
  & \hspace*{-3.8ex}\rotatebox{60}{\scalebox{0.8}{\parbox{4em}{\centering\footnotesize VisualBERT Google}}}
  & \hspace*{-3.8ex}\rotatebox{60}{\scalebox{0.8}{\parbox{4em}{\centering\footnotesize ViLBERT Google}}}
  & \hspace*{-3.8ex}\rotatebox{60}{\scalebox{0.8}{\parbox{4em}{\centering\footnotesize LXMert Google}}}
  & \hspace*{-3.8ex}\rotatebox{60}{\scalebox{0.8}{\parbox{4em}{\centering\footnotesize VLBERT Google}}}
  
  & \hspace*{-3.8ex}\rotatebox{60}{\scalebox{0.8}{\parbox{4em}{\centering\footnotesize VisualBERT Google}}}
  & \hspace*{-3.8ex}\rotatebox{60}{\scalebox{0.8}{\parbox{4em}{\centering\footnotesize ViLBERT Google}}}
  & \hspace*{-3.8ex}\rotatebox{60}{\scalebox{0.8}{\parbox{4em}{\centering\footnotesize LXMert Google}}}
  & \hspace*{-3.8ex}\rotatebox{60}{\scalebox{0.8}{\parbox{4em}{\centering\footnotesize VLBERT Google}}}

  & \hspace*{-3.8ex}\rotatebox{60}{\scalebox{0.8}{\parbox{4em}{\centering\footnotesize Mask}}}
  & \hspace*{-3.8ex}\rotatebox{60}{\scalebox{0.8}{\parbox{4em}{\centering\footnotesize ViLBERT Google}}}
  & \hspace*{-3.8ex}\rotatebox{60}{\scalebox{0.8}{\parbox{4em}{\centering\footnotesize ViLBERT Google}}}
  & \hspace*{-3.8ex}\rotatebox{60}{\scalebox{0.8}{\parbox{4em}{\centering\footnotesize LXMert Google}}}
  & \hspace*{-3.8ex}\rotatebox{60}{\scalebox{0.8}{\parbox{4em}{\centering\footnotesize VLBERT Google}}}
    \vspace{-3ex}\\%
  & W%
  & 2%
  & 4%
  & 4%
  & 3%
  & 3%
  & 4%
  & 5%
  & 4%
  &T%
  &- %
  & 0%
  & 5%
  & 2%
  \\%
  & {\leavevmode\hspace*{0.4ex}S}%
  & 3%
  & 4%
  & 5%
  & 3%
  & 4%
  & 3%
  & 5%
  & 5%
  &I%
  & -%
  & 4%
  & 5%
  & 3%
  \\%
  & {\leavevmode\hspace*{0.2ex}C}%
  & 5%
  & 7%
  & 5%
  & 6%
  & 6%
  & 4%
  & 5%
  & 6%
  & %
  & %
  & 
  &
  \\%
\end{tabular*}}%

\caption{A summary of all previous results on the new image dataset derived from
  Google searches showing the number of significant bias test partitioned by the
  type of test. There are a total of 6 gender bias tests and 7 race bias
  test. Experiments 1 and 2 show no strong differences between models while in
  Experiment 3 ViLBERT stands out.}
\end{table*}

Overall, the results are consistent with prior work on biases in both humans and with ungrounded
models such as BERT.
%
%
Following~\citet{tan2019assessing}, each experiment examines
the bias in three types of embeddings: word embeddings,
sentence embeddings, and contextualized word embeddings.
While there is broad agreement between these different ways of using embeddings,
they are not identical in terms of which biases are discovered.
It is unclear which of these methods is more sensitive, and which finds
biases that are more consequential in predicting the results of a larger system
constructed from these models.
Methods to mitigate biases will hopefully address all three embedding types and
all of the three questions we restate below.

\textbf{Do joint embeddings encode social biases?} See Experiment 1,
\cref{sec:exp1}.
The results presented in~\cref{tab:exp1} and~\cref{tab:expC} clearly indicate that
the answer is yes.
Numerous biases are uncovered with results that are broadly compatible with
\citet{may2019measuringbias} and \citet{tan2019assessing}.
It appears that more pronounced social biases exist in  grounded compared to
ungrounded embeddings.

\textbf{Can grounded evidence that counters a stereotype alleviate biases?} See Experiment 2, \cref{sec:exp2}.
The results presented in~\cref{tab:exp2} and~\cref{tab:expC} indicate that the answer is no.
Biases are somewhat attenuated when models are shown evidence against them, but
overall, preconceptions about biases tend to overrule direct visual evidence to
the contrary.
This is worrisome for the applications of such models.
In particular, using such models to search or filter data in the
service of creating new datasets may well introduce new biases.

\textbf{To what degree are encoded biases in joint embeddings from language or vision?} See Experiment 3, \cref{sec:exp3}.
The results for the second variant of Experiment 3 which is performed by masking the input text or image are presented in~\cref{tab:exp3} and~\cref{tab:expC} are generally
significant, more so for language than vision.
We report results for the sentence-level encoding and observed similar results for the word-level encoding.
We did not measure contextual encodings as they would include the encoding for the \texttt{[MASK]} token.
This indicates that biases arise from both modalities, but this does differ by model architecture.
For VL-BERT language appears to dominate.  
The results for the first variant of Experiment 3 congruent with these results, with, large effect sizes ($s$=0.42 for ViLBERT and $s$=0.467 for VisualBERT with 12\% of tests being statistically significant) demonstrating that language contributes more than vision.
It could be that the biases in
language are so powerful that vision does not contribute to them given that in
any one example it appears unable to override the existing biases (experiment
2).
It is encouraging that models do consider vision, but the
differing biases in vision and text do not appear to
help.

\section{Discussion}

Visually grounded embeddings have biases similar to ungrounded embeddings and vision does not appear to help eliminate
them.
At test time, vision has difficulty overcoming biases, even when presented counter-stereotypical evidence.
This is worrisome for deployed systems that use such embeddings, as it indicates that
they ignore visual evidence that a bias does not hold for a particular
interaction.
Overall, language and vision each contribute to encoded bias, yet the means of using vision to mitigate is not immediately clear.
We enumerated the combinations of inputs possible in the grounded setting and
selected three interpretable questions that we answered above.
Other questions could potentially be asked using the dataset we developed,
although we did not find any others that were intuitive or
non-redundant.

While we discuss joint vision and language embeddings, the methods introduced
here apply to any grounded embeddings, such as joint audio and language
embeddings~\citep{kiela2015multi,torabi2016learning}.
Measuring bias in such data would require collecting a new dataset, but could
use our metrics, Grounded-WEAT and Grounded-SEAT, to answer the same three
questions.

Many joint models are transferred to a new dataset without fine-tuning.
We demonstrate that going out-of-domain into a new dataset amplifies biases.
This need not be so: out-of-domain models have worse performance which might
result in fewer biases.
We did not test task-specific fine-tuned models, but intend to do so in the
future.

Humans clearly have biases, not just machines.
Although, initial evidence indicates that when faced with examples that go
against prejudices, \ie counter-stereotyping, there is a significant reduction
in human biases~\citep{peck2013putting,columb2016obama}.
Straightforward applications of this idea are far from trivial, as
\citet{wang2019balanced} show that merely balancing a dataset by a certain
attribute is not enough to eliminate bias.
Perhaps artificially manipulating visual datasets can debias shared embeddings.
We hope that these datasets and metrics will lead to understanding human
biases in grounded settings as well as the development of new methods to
debias representations.

%

\section*{Acknowledgments}

This work was supported by the Center for Brains, Minds and Machines, NSF STC
award 1231216, the Toyota Research Institute, the MIT CSAIL Systems that Learn
Initiative, the NSF Graduate Research Fellowship, the DARPA GAILA program, the United States Air Force Research Laboratory and
United States Air Force Artificial Intelligence Accelerator under Cooperative Agreement Number
FA8750-19-2-1000, and the Office of Naval Research under Award Number
N00014-20-1-2589 and Award Number N00014-20-1-2643. The views and conclusions
contained in this document are those of the authors and should not be
interpreted as representing the official policies, either expressed or implied,
of the U.S. Government. The U.S. Government is authorized to reproduce and
distribute reprints for Government purposes notwithstanding any copyright
notation herein.

\section*{Ethical Considerations}

We would like to urge subsequent work to avoid a common ethical problem we have noticed while reviewing the literature on bias in NLP. Much prior work refers to gender as ``male'' and ``female'', thereby conflating gender and sex. Recent work in psychology has disentangled these two concepts, and conflating them both blinds us to a type of bias while actively causing harm.

Our approach studies societal biases in models.
These biases are inherently unjust, predisposing models toward judging people by skin color, age, etc.
They are also practically damaging; they can result in real-world consequences.
As part of large systems these biases may not be apparent as the source of discrimination, and it may not even be apparent that systems are treating individuals differently.
People may even acclimatize to being treated differently or may interpret a machine discriminating based on race or gender as an inevitable but fair consequence of using a particular algorithm.
We vehemently disagree.
All systems and algorithm choices are made by humans, all data is curated by humans, and ultimately humans decide what to do with and when to use models.
All unequal outcomes are a deliberate choice; engineers should not be able to hide behind the excuse of a black-box or a complex algorithm.
We believe that by revealing biases, by providing tests for biases that are as focused as possible on the smallest units of systems, we can both assist the development of better models and allow the auditing of models to ascertain their fairness.

Data was collected in an ethical manner approved by the institution IRB board. No crowdsourced workers were employed. Instead we used a \emph{top k} keyword search on Google Images.
Because we collected images from the web, there is no straightforward way to use self-identified characteristics for gender and race.
We expect biases and preconceived notions of identity to have some bearing on label accuracy.
The dataset includes images available for free on the web and simple captions, \eg Here is a man.

The biases we evaluate in this paper are based on various theories and works in psychology, such as the trope of the angry Black woman.
Of course, that literature itself is limited; there are many biases which affect billions of people but do not appear in any available test, e.g., for almost any ethnic group there are those who will believe they do not work hard, but there are virtually no ethnic-group-specific tests.
There are also likely biases which we have not yet articulated.
Unfortunately, at present there is no coherent theory of biases to generate an exhaustive list and test them.

\bibliography{anthology,custom}

\begin{thebibliography}{27}
\expandafter\ifx\csname natexlab\endcsname\relax\def\natexlab#1{#1}\fi

\bibitem[{Bertrand and Mullainathan(2004)}]{bertrand2004emily}
Marianne Bertrand and Sendhil Mullainathan. 2004.
\newblock Are emily and greg more employable than lakisha and jamal? a field
  experiment on labor market discrimination.
\newblock \emph{American Economic Review}, 94(4):991--1013.

\bibitem[{Blodgett et~al.(2020)Blodgett, Barocas, Daumé~III, and
  Wallach}]{blodgett2020survey}
Su~Lin Blodgett, Solon Barocas, Hal Daumé~III, and Hann Wallach. 2020.
\newblock Language (technology) is power: A critical survey of “bias” in
  nlp.
\newblock In \emph{Proceedings of ACL}.

\bibitem[{Caliskan et~al.(2017)Caliskan, Bryson, and
  Narayanan}]{caliskan2017semantics}
Aylin Caliskan, Joanna~J Bryson, and Arvind Narayanan. 2017.
\newblock Semantics derived automatically from language corpora contain
  human-like biases.
\newblock \emph{Science}, 356(6334):183--186.

\bibitem[{Chen et~al.(2015)Chen, Fang, Lin, Vedantam, Gupta, Doll{\'a}r, and
  Zitnick}]{chen2015microsoft}
Xinlei Chen, Hao Fang, Tsung-Yi Lin, Ramakrishna Vedantam, Saurabh Gupta, Piotr
  Doll{\'a}r, and C~Lawrence Zitnick. 2015.
\newblock Microsoft coco captions: Data collection and evaluation server.
\newblock \emph{arXiv:1504.00325}.

\bibitem[{Collins(2004)}]{collins2004black}
Patricia~Hill Collins. 2004.
\newblock \emph{Black sexual politics: African Americans, gender, and the new
  racism}.
\newblock Routledge.

\bibitem[{Columb and Plant(2016)}]{columb2016obama}
Corey Columb and E~Ashby Plant. 2016.
\newblock The obama effect six years later: The effect of exposure to obama on
  implicit anti-black evaluative bias and implicit racial stereotyping.
\newblock \emph{Social Cognition}, 34(6):523--543.

\bibitem[{Devlin et~al.(2018)Devlin, Chang, Lee, and
  Toutanova}]{devlin2018bert}
Jacob Devlin, Ming-Wei Chang, Kenton Lee, and Kristina Toutanova. 2018.
\newblock {BERT}: Pre-training of deep bidirectional transformers for language
  understanding.
\newblock \emph{arXiv:1810.04805}.

\bibitem[{Drozd et~al.(2016)Drozd, Gladkova, and Matsuoka}]{drozd2016word}
Aleksandr Drozd, Anna Gladkova, and Satoshi Matsuoka. 2016.
\newblock Word embeddings, analogies, and machine learning: Beyond king-man+
  woman= queen.
\newblock In \emph{Proceedings of coling 2016, the 26th international
  conference on computational linguistics: Technical papers}, pages 3519--3530.

\bibitem[{Greenwald et~al.(1998)Greenwald, McGhee, and
  Schwartz}]{greenwald1998measuring}
Anthony~G Greenwald, Debbie~E McGhee, and Jordan~LK Schwartz. 1998.
\newblock Measuring individual differences in implicit cognition: the implicit
  association test.
\newblock \emph{Journal of personality and social psychology}, 74(6):1464.

\bibitem[{Hu et~al.(2019)Hu, Singh, Darrell, and Rohrbach}]{hu2019iterative}
Ronghang Hu, Amanpreet Singh, Trevor Darrell, and Marcus Rohrbach. 2019.
\newblock Iterative answer prediction with pointer-augmented multimodal
  transformers for textvqa.
\newblock \emph{arXiv:1911.06258}.

\bibitem[{Kiela and Clark(2015)}]{kiela2015multi}
Douwe Kiela and Stephen Clark. 2015.
\newblock Multi-and cross-modal semantics beyond vision: Grounding in auditory
  perception.
\newblock In \emph{Proceedings of the 2015 Conference on Empirical Methods in
  Natural Language Processing}, pages 2461--2470.

\bibitem[{Li et~al.(2019)Li, Yatskar, Yin, Hsieh, and Chang}]{li2019visualbert}
Liunian~Harold Li, Mark Yatskar, Da~Yin, Cho-Jui Hsieh, and Kai-Wei Chang.
  2019.
\newblock {VisualBERT}: A simple and performant baseline for vision and
  language.
\newblock \emph{arXiv:1908.03557}.

\bibitem[{Lu et~al.(2019)Lu, Batra, Parikh, and Lee}]{lu2019vilbert}
Jiasen Lu, Dhruv Batra, Devi Parikh, and Stefan Lee. 2019.
\newblock {ViLBERT}: Pretraining task-agnostic visiolinguistic representations
  for vision-and-language tasks.
\newblock In \emph{Advances in Neural Information Processing Systems}, pages
  13--23.

\bibitem[{May et~al.(2019)May, Wang, Bordia, Bowman, and
  Rudinger}]{may2019measuringbias}
Chandler May, Alex Wang, Shikha Bordia, Samuel~R Bowman, and Rachel Rudinger.
  2019.
\newblock On measuring social biases in sentence encoders.
\newblock \emph{arXiv:1903.10561}.

\bibitem[{Mikolov et~al.(2013)Mikolov, Sutskever, Chen, Corrado, and
  Dean}]{mikolov2013distributed}
Tomas Mikolov, Ilya Sutskever, Kai Chen, Greg~S Corrado, and Jeff Dean. 2013.
\newblock Distributed representations of words and phrases and their
  compositionality.
\newblock In \emph{Advances in neural information processing systems}, pages
  3111--3119.

\bibitem[{Murahari et~al.(2019)Murahari, Batra, Parikh, and
  Das}]{murahari2019large}
Vishvak Murahari, Dhruv Batra, Devi Parikh, and Abhishek Das. 2019.
\newblock Large-scale pretraining for visual dialog: A simple state-of-the-art
  baseline.
\newblock \emph{arXiv:1912.02379}.

\bibitem[{Peck et~al.(2013)Peck, Seinfeld, Aglioti, and
  Slater}]{peck2013putting}
Tabitha~C Peck, Sofia Seinfeld, Salvatore~M Aglioti, and Mel Slater. 2013.
\newblock Putting yourself in the skin of a black avatar reduces implicit
  racial bias.
\newblock \emph{Consciousness and cognition}, 22(3):779--787.

\bibitem[{Peters et~al.(2018)Peters, Neumann, Iyyer, Gardner, Clark, Lee, and
  Zettlemoyer}]{peters2018deep}
Matthew~E Peters, Mark Neumann, Mohit Iyyer, Matt Gardner, Christopher Clark,
  Kenton Lee, and Luke Zettlemoyer. 2018.
\newblock Deep contextualized word representations.
\newblock \emph{arXiv:1802.05365}.

\bibitem[{Radford et~al.(2018)Radford, Narasimhan, Salimans, and
  Sutskever}]{radford2018GPT}
Alec Radford, Karthik Narasimhan, Tim Salimans, and Ilya Sutskever. 2018.
\newblock Improving language understanding by generative pre-training.
\newblock \emph{OpenAI preprint}.

\bibitem[{Sharma et~al.(2018)Sharma, Ding, Goodman, and
  Soricut}]{sharma2018conceptual}
Piyush Sharma, Nan Ding, Sebastian Goodman, and Radu Soricut. 2018.
\newblock Conceptual captions: A cleaned, hypernymed, image alt-text dataset
  for automatic image captioning.
\newblock In \emph{Proceedings of ACL}.

\bibitem[{Su et~al.(2019)Su, Zhu, Cao, Li, Lu, Wei, and Dai}]{su2019vl}
Weijie Su, Xizhou Zhu, Yue Cao, Bin Li, Lewei Lu, Furu Wei, and Jifeng Dai.
  2019.
\newblock Vl-bert: Pre-training of generic visual-linguistic representations.
\newblock \emph{arXiv preprint arXiv:1908.08530}.

\bibitem[{Tan and Bansal(2019)}]{tan2019lxmert}
Hao Tan and Mohit Bansal. 2019.
\newblock Lxmert: Learning cross-modality encoder representations from
  transformers.
\newblock \emph{arXiv preprint arXiv:1908.07490}.

\bibitem[{Tan and Celis(2019)}]{tan2019assessing}
Yi~Chern Tan and L~Elisa Celis. 2019.
\newblock Assessing social and intersectional biases in contextualized word
  representations.
\newblock In \emph{Advances in Neural Information Processing Systems}, pages
  13209--13220.

\bibitem[{Torabi et~al.(2016)Torabi, Tandon, and Sigal}]{torabi2016learning}
Atousa Torabi, Niket Tandon, and Leonid Sigal. 2016.
\newblock Learning language-visual embedding for movie understanding with
  natural-language.
\newblock \emph{arXiv:1609.08124}.

\bibitem[{Wang et~al.(2019)Wang, Zhao, Yatskar, Chang, and
  Ordonez}]{wang2019balanced}
Tianlu Wang, Jieyu Zhao, Mark Yatskar, Kai-Wei Chang, and Vicente Ordonez.
  2019.
\newblock Balanced datasets are not enough: Estimating and mitigating gender
  bias in deep image representations.
\newblock In \emph{Proceedings of the IEEE International Conference on Computer
  Vision}, pages 5310--5319.

\bibitem[{Zhao et~al.(2019)Zhao, Wang, Yatskar, Cotterell, Ordonez, and
  Chang}]{zhao2019gender}
Jieyu Zhao, Tianlu Wang, Mark Yatskar, Ryan Cotterell, Vicente Ordonez, and
  Kai-Wei Chang. 2019.
\newblock Gender bias in contextualized word embeddings.
\newblock \emph{arXiv:1904.03310}.

\bibitem[{Zhao et~al.(2018)Zhao, Wang, Yatskar, Ordonez, and
  Chang}]{zhao2018gender}
Jieyu Zhao, Tianlu Wang, Mark Yatskar, Vicente Ordonez, and Kai-Wei Chang.
  2018.
\newblock Gender bias in coreference resolution: Evaluation and debiasing
  methods.
\newblock \emph{arXiv:1804.06876}.

\end{thebibliography}
\bibliographystyle{acl_natbib}




\end{document}